\documentclass{article}

\usepackage{microtype}
\usepackage{graphicx}
\usepackage{subfigure}
\usepackage{booktabs} % for professional tables
\usepackage{placeins}
\usepackage[many]{tcolorbox}
\usepackage{subfigure}
\usepackage{hyperref}

\usepackage[accepted]{icml2025}

\usepackage{amsmath}
\usepackage{amssymb}
\usepackage{mathtools}
\usepackage{amsthm}
\usepackage{enumitem}

\usepackage[capitalize,noabbrev]{cleveref}

\theoremstyle{plain}

\theoremstyle{definition}

\theoremstyle{remark}

\usepackage[textsize=tiny]{todonotes}
\usepackage{subcaption}
\icmltitlerunning{Multiscale Byte Language Models}

\begin{document}

\twocolumn[
    \icmltitle{\textsc{Multiscale Byte Language Models} \\
               A Hierarchical Architecture for Causal Million-Length Sequence Modeling}
    
    
    
    
    % \begin{icmlauthorlist}
    % \icmlauthor{Eric Egli}{yyy}
    % \icmlauthor{Matteo Manica}{yyy}
    % \icmlauthor{Jannis Born}{yyy}
    % \end{icmlauthorlist}

    \begin{icmlauthorlist}
    \icmlauthor{Eric Egli,}{}
    \icmlauthor{Matteo Manica,}{}
    \icmlauthor{Jannis Born}{}
    \end{icmlauthorlist}
    
    \vskip -0.2in
    \begin{center}
        IBM Research Europe \\
        % \texttt{jab@zurich.ibm.com}
    \end{center}

    \vskip 0.3in
    
    % \icmlaffiliation{yyy}{IBM Research Europe}
    % % \icmlaffiliation{comp}{Company Name, Location, Country}
    % % \icmlaffiliation{sch}{School of ZZZ, Institute of WWW, Location, Country}
    
    % \icmlcorrespondingauthor{Eric Egli}{eric.christian.egli@ibm.com}
    % \icmlcorrespondingauthor{Jannis Born}{jab@zurich.ibm.com}
    % \icmlcorrespondingauthor{Matteo Manica}{tte@zurich.ibm.com}
    
    \icmlkeywords{Machine Learning, Byte Language Models, Tokenization-free methods, Mamba, Transformers, Context Length, Multimodal, VQA}
    
    \vskip 0.3in
]


\newcommand{\TODO}[1]{{\color{red}#1}}
\newcommand{\R}{\mathbb{R}}
\newcommand{\mbf}[1]{\mathbf{#1}}
\newcommand{\ov}[1]{\overline{#1}}
\newcommand{\bs}[1]{\boldsymbol{#1}}


\printAffiliationsAndNotice{}  % leave blank if no need to mention equal contribution

\begin{abstract}
Bytes form the basis of the digital world and thus are a promising building block for multimodal foundation models. 
Recently, Byte Language Models (BLMs) have emerged to overcome tokenization, yet the excessive length of bytestreams requires new architectural paradigms.
Therefore, we present the Multiscale Byte Language Model (MBLM), a model-agnostic hierarchical decoder stack that allows training with context windows of $5$M bytes on single GPU in full model precision. We thoroughly examine MBLM's performance with Transformer and Mamba blocks on both unimodal and multimodal tasks. Our experiments demonstrate that hybrid architectures are efficient in handling extremely long byte sequences during training while achieving near-linear generational efficiency. 
To the best of our knowledge, we present the first evaluation of BLMs on visual Q\&A tasks and find that, despite serializing images and the absence of an encoder, a MBLM with pure next token prediction can match custom CNN-LSTM architectures with designated classification heads. 
We show that MBLMs exhibit strong adaptability in integrating diverse data representations, including pixel and image filestream bytes, underlining their potential toward omnimodal foundation models.
Source code is publicly available at:~\url{https://github.com/ai4sd/multiscale-byte-lm}.
% with universal LM head operating on non-spatial (i.e., serialized) data can 
% The BLMs can seamlessly leverage information from different representations such as pixel bytes or image filestream bytes.
\end{abstract}

\section{Introduction}
From the perspective of a traditional computational linguist, the success of Language Models (LMs) in NLP is a success story of replacing inductive biases with data-agnostic computation blocks. 
Yet, (sub)-word tokenization has remained a cornerstone of any LM workflow, inducing strong assumptions about the structure of text and hampering out-of-distribution learning. 
In contrast, tokenization-free models reduce the preprocessing overhead \cite{byt5tokenfree} by utilizing bytes as a universal encoding format and enable seamless adaptation to diverse languages and modalities \cite{bytelevelmachinetrans,li2019bytes}. 
Additionally, byte-language models (BLMs) pre-trained on mixed-modality datasets exhibit performance comparable to models trained on specific modalities \cite{digitalworldsimulators}. 
To address the challenges of long sequences and the computational overhead associated with byte-level granularity, prior work has aimed to mitigate the quadratic complexity of autoregressive Transformers with computationally more efficient, hierarchical Transformers \cite{megabyte, pagnoni2024bytelatenttransformerpatches} or Mamba models optimized for fast inference \cite{mambabyte}. However, these approaches depend on modality-specific optimizations and model-specific features, which limit their generalization and scalability.\\
\begin{figure}[!htb]
    \centering
    \includegraphics[width=0.85\linewidth]{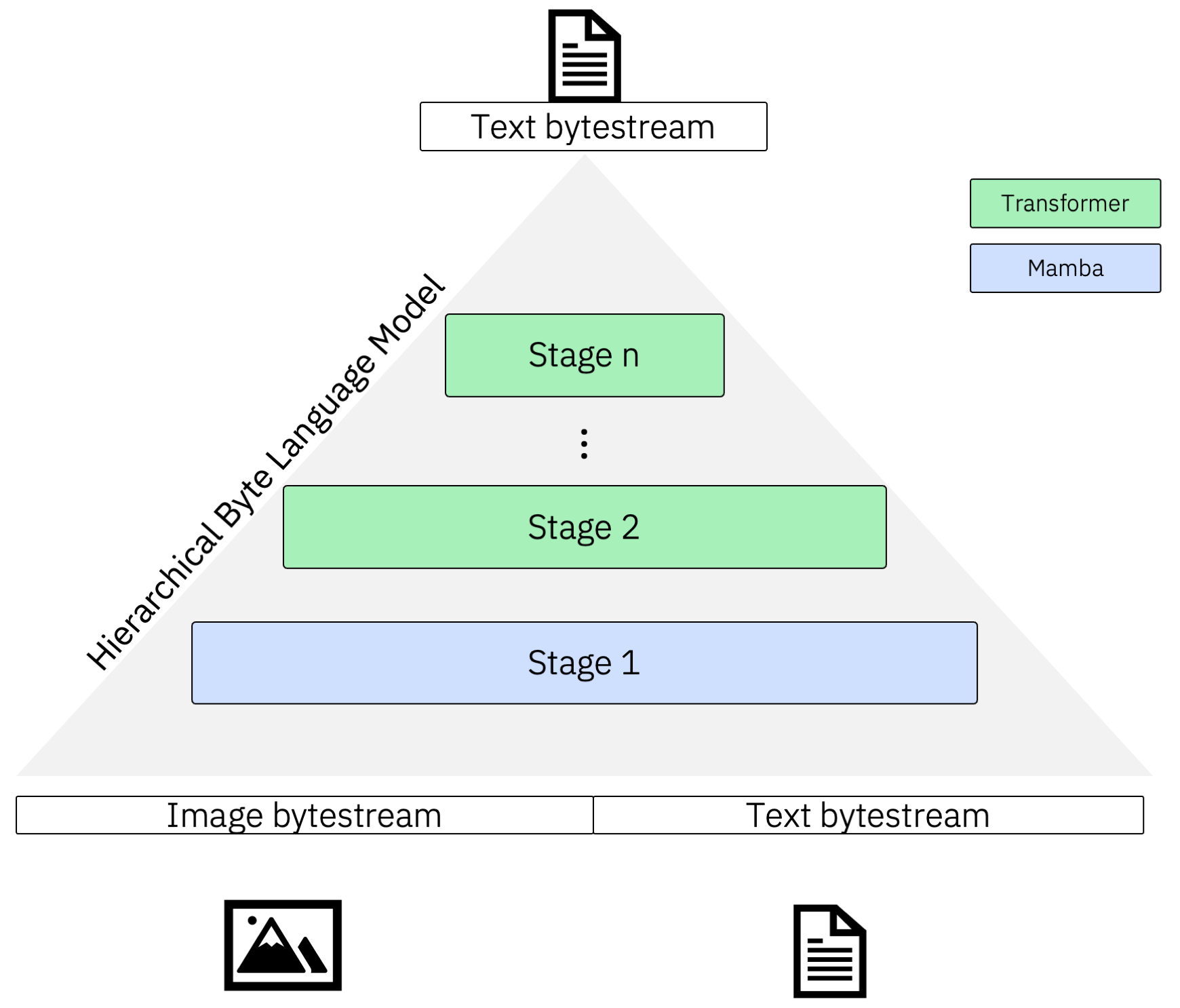}
    \caption{
    \textbf{The Multiscale Byte Language Model (MBLM)} processes bytestreams from any modality that can be serialized into bytes. Each stage in the hierarchical architecture employs a decoder model to generate a new representation for input patches, which is subsequently passed to the next stage as augmented input. The final output of the MBLM is a bytestream formed by concatenating the outputs of the last stage, $n$.}
    \label{fig:mblm}
\end{figure}
As shown in~\autoref{fig:mblm}, we here introduce the Multiscale Byte Language Model (MBLM), a model-- and modality-agnostic architecture for causal byte language modeling. MBLMs extend the MegaByte hierarchy \cite{megabyte} to an unlimited number of stages, and predict the next byte of a large input bytestream by refining input sequence representations through a hierarchy of generic decoder models, while enabling precise control over stage parallelism.
By integrating Transformer decoders \cite{attnisallyouneed} with Mamba \cite{mambabyte, mamba2}, we demonstrate that hybrid hierarchies minimize computational requirements during both training and inference, outperforming other architectures when handling input sequences comprising hundreds of thousands of bytes. MBLMs provide granular control over the trade-off between parallelism and compute time by selectively checkpointing intermediate activations, enabling efficient training on sequences spanning up to 5M bytes on single GPU. Our approach supports modality-independent pre-training and fine-tuning on multimodal downstream tasks. 
In a novel application of byte models to multimodal input data, we achieve performance comparable to a robust multimodal baseline on visual Q\&A tasks using only a language modeling head.   

\section{Related work}
The MBLM builds upon the design principles of MegaByte \cite{megabyte}, a causal byte language model featuring a hierarchical architecture of two Transformer decoders, enabling subquadratic self-attention and context windows up to 1.2M bytes. MegaByte processes patch representations of the input sequence with a global decoder, refines these representations, and feeds them into a local model that autoregressively predicts individual bytes. Incorporating the Mamba architecture \cite{mamba, mamba2} at the byte level, MambaByte \cite{mambabyte} demonstrated superior performance over MegaByte in a FLOP-controlled setting across various datasets. As an alternative to the fixed-size patching used in MegaByte, \citet{pagnoni2024bytelatenttransformerpatches} proposed the Byte Latent Transformer (BLT), which dynamically segments bytes into patches based on the entropy of the next byte. 
BLT demonstrated that byte language models can be efficiently scaled, achieving performance comparable to a subword-based LLama 3 model \cite{grattafiori2024llama3herdmodels} at the 8B parameter scale.\\
However, none of these approaches have demonstrated the capability to handle multimodal inputs, which is arguably the most inherent strength of byte-level models. As shown by \citet{digitalworldsimulators} with bGPT, extending pre-training to include binary data from mixed modalities facilitates effective cross-modality knowledge transfer. This reinforces the hypothesis that byte-level models uniquely capture features and patterns in ubiquitous bytestreams, irrespective of the original data format. Nevertheless, limited focus has been placed on architectures capable of translating multimodal inputs into multimodal outputs. Perceiver IO \cite{jaegle2022perceiverio} addresses this by mapping inputs of arbitrary size into a latent space using a latent array that encodes the semantics of the input. The latent representation is iteratively refined through a series of attention modules and subsequently decoded into outputs of arbitrary shape via an output query array. Due to the encoder and decoder attention modules scaling linearly with the input and output size, and most of the computation occurring in the latent attention modules, Perceiver IO can efficiently handle extremely large input and output dimensions.
Yet, PerceiverIO explores bytes only to represent text and thus, to date, we still lack applications of BLMs on multimodal tasks like visual Q\&A.

\section{Methods}
\subsection{MBLM}
The MBLM \emph{module} consists of $N$ causal decoder \emph{models} 
$M_{i\leq N}$ that are stacked hierarchically. The first $N-1$ stages $M_1, \ldots, M_{N-1}$ contain \textbf{global models}, while the final stage $M_{N}$ contains the \textbf{local model}. Each model $M_i$ operates on inputs with a hidden state of dimension $D_i\in\R$ and a patch/context size $P_i\in\R$. Inputs to an MBLM module are sequences of $B$ batches, each of length $L$. The vocabulary $V$ consists of 256 tokens for byte-level settings. Similar to MegaByte \cite{megabyte}, MBLMs scale through input length compression and aim to operate on sequences of length $L_{\max}=\prod_{i=1}^NP_i$. 

\subsubsection*{Patch Embedder}
MBLMs employ a patch embedder that inputs a discrete sequence $\mbf{x} \in \R^{B \times L}$, embeds each element, adds positional encodings and then chunks it into patches for each stage.
\begin{enumerate}[wide, labelwidth=!, labelindent=0pt]
    \item \textbf{Embed} the bytes in $\mbf{x}$ for each stage $i$:
    \begin{equation}
        \mbf{x}^\text{emb}_i \in \R^{B \times L \times D_N} =E_i^\text{emb}(\mbf{x}) + E_i^\text{pos}(\mbf{x})
    \end{equation}
    Although we always use $D_N$ as the embedding dimension, each model learns its own byte embedding. $E_i^\text{emb}\in\R^{V \times D_N}$ and $E_i^\text{pos}\in\R^{P_i \times D_N}$ are thus specific to each stage.
    \item \textbf{Reshape} $\mathbf{x}^\text{emb}$ to a nested sequence of patch embeddings $\mathcal{P}^\text{emb}$ for all $i$:
    \begin{equation}
        \mathbf{x}_i^\text{emb}\xrightarrow{\text{reshape}}\mathcal{P}_i^\text{emb} \in \R^{B \times P_1 \times \ldots \times P_N \times D_N}
    \end{equation}
    If $L$ cannot be factored into $P_1 \times \ldots \times P_N$, the inner sequence lengths $P_2$ to $P_N$ are padded with a padding token. In case $L>L_{\max}$ and no positional embeddings are applied, we additionally allow $P_1$ to be larger than specified, enabling MBLMs to operate on longer inputs by extending the first global model's context window. 
    \item \textbf{Project token embeddings to patches} for the global stages. Recall that all $\mathcal{P}^\text{emb}_i \in \R^{B \times P_1 \times \ldots \times P_N \times D_N}$ are of the same shape. For each stage, we flatten the embeddings and apply a linear projection to the model dimension of stage $i$:
    \begin{equation}
    \begin{split}
        \boldsymbol{W}_i^{\text{patch}}:\quad&\R^{B \times P_1 \times \ldots \times P_i \times (\bs{P_{i+1}} \times \ldots \times \bs{P_N} \times \bs{D_N})}\\ \rightarrow\quad &\R^{B \times P_1 \times \ldots \times \bs{P_i} \times \bs{D_i}}
    \end{split}
    \end{equation}
    We furthermore prepend a trainable start token $E^\text{pad}_i\in\R^{D_i}$ to the start of each patch $P_i$ and drop the last patch from the projection to match lengths:
    \begin{equation}
    \begin{split}
       \mathcal{P}_i^\text{emb}\in\R^{B \times P_1 \ldots \times P_i \times D_i} &=\\ \text{concat}(E^\text{pad}_i, \mathcal{P}_i^\text{emb}\bs{W}_i^{\text{patch}})
    \end{split}
    \end{equation}
\end{enumerate}
\subsubsection*{Global Model Projections}
The global models perform \textbf{inter-patch} modeling by capturing dependencies between patches and output updated patch representations. These updated representations are added to the token embeddings of the next stage, allowing patches to receive global sequence information from the leftward context. In order to process all patches contained in $\mathcal{P}_i^\text{emb}$ in parallel with $M_i$, we reshape $\mathcal{P}_i^\text{emb}$ to a new batch dimension $K_i$:
\begin{equation}
    \mathcal{P}_i^\text{emb}\in\R^{B \times P_1 \ldots \times P_i \times D_i} \xrightarrow{\text{pack}} \mathcal{P}_i^{\text{emb}^\mbf{\prime}}\in\R^{\bs{K_i} \times P_i \times D_i}
\end{equation}
\begin{equation*}
    \text{with }K_i=B \cdot \displaystyle \prod_{j=1}^{i-1}P_j\quad\forall i>1
\end{equation*}
For deep hierarchies, $K\in\R$ can become very large. For this reason, we let all but the first stage trade performance for memory efficiency by leveraging gradient checkpointing. Instead of processing all $K$ patches in parallel, we \emph{optionally} divide them into $c$ smaller chunks that are processed sequentially and recompute intermediate activations during the backward pass. This approach allows for much larger batch sizes and input sequences to fit within memory constraints, albeit at the cost of increased computation time during training. To propagate information to higher stages, outputs of global stage $i$ are linearly projected to the dimension of the next stage $i+1$ with $\bs{W}_i^{\text{global}}: \R^{D_i} \rightarrow \R^{D_{i+1}}$ and added to the patch embedding $\mathcal{P}^{\text{emb}^\prime}$ of the next stage. The projection is offset so that the trainable start tokens do not receive patch representations. Expressed as a recurrence relation with $M_i$ being the model at stage $i$:
\begin{align}
    \overbrace{\mathcal{P}_{i}^\text{in}}^{\text{Input to $M_i$}}&=\overbrace{\mathcal{P}_{i}^{\text{emb}^\prime}}^{\text{Embedding of stage $i$}}+\overbrace{\mathcal{P}_{i-1}^\text{out}\bs{W}_{i-1}^{\text{global}}}^{\text{Output of $M_{i-1}$}}\\
    \overbrace{\mathcal{P}_{i}^\text{out}}^{\text{Output of $M_i$}}&=\underset{c}{\text{concat}}\left(M_i(\mathcal{P}_i^\text{in})\right)
\end{align}
\begin{equation*}
    \text{with }\mathcal{P}_i^\text{in}\in\R^{\frac{K_i}{c} \times P_i \times D_i}
\end{equation*}
For the first global stage, there is no parent patch representation. The base case of the recurrence and input to the first model in the hierarchy is thus given by $\mathcal{P}_{1}^\text{in}=\mathcal{P}_{1}^{\text{emb}^\prime}\in\R^{B \times P_1 \times D_1}$. Importantly, at any stage, the inputs and outputs are of the same shape $\in\R^{K_i \times P_i \times D_i}$. In essence, this is what makes MBLMs model-agnostic: Any model that implements a sequence transformation to an equally sized output qualifies as a stage model. However, because MBLMs are designed for \emph{causal} byte-language modeling, the stage models need to be autoregressive, since they are only given information from the left context.

\subsubsection*{Local intra-patch modeling}
The input to the local stage is given by $\mathcal{P}_{N}^\text{in}\in\R^{K_N \times P_N \times D_N}$. Unlike the global models, whose primary role is to contextualize patches, the local model performs byte-level \textbf{intra-patch} modeling by autoregressively predicting individual bytes starting from the trainable start token located at $P_{N, 0}$. We provide the same parallelism tradeoff via gradient checkpointing as for the global models. The output of the local model is then projected to logits $Z$ through a linear layer:
\begin{equation}
    Z\in\R^{K_N \times P_N \times V}=\mathcal{P}_{N}^\text{out}\bs{W}^{\text{head}}
\end{equation}
Afterward, $Z$ is reshaped into $\mathbf{y} \in \mathbb{R}^{B \times L \times V}$. All but the initial start token within the local patches are removed, and the cross-entropy loss is calculated over the final sequence to train the model for next-token prediction.
For an exemplary visualization of a 3D MBLM see~\autoref{fig:mblm-simple}.

\begin{figure}[!htb]
    \centering
    \includegraphics[width=\linewidth]{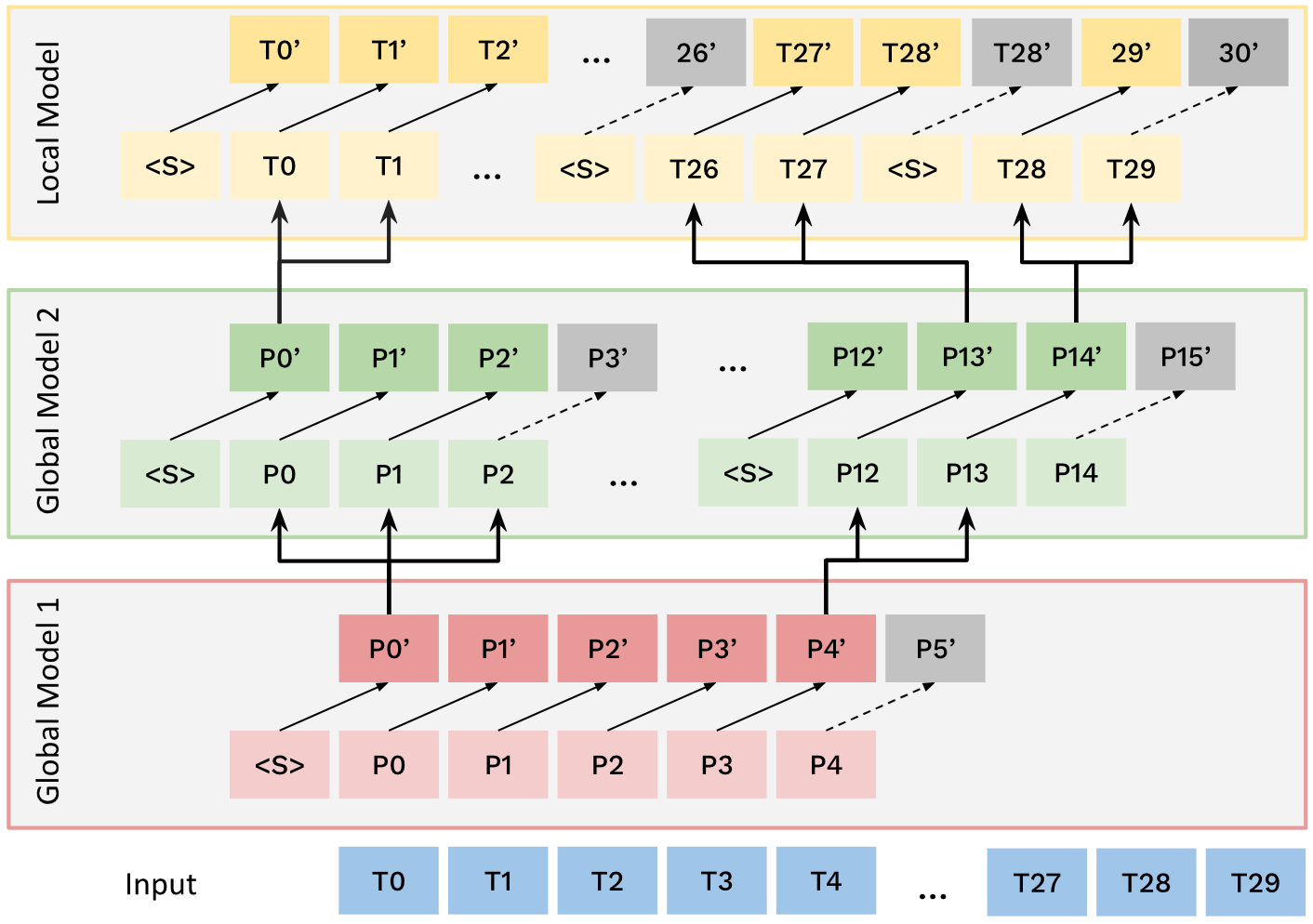}
    \caption{
   A 3D MBLM module with two global and one local decoder models and corresponding patch sizes $P_1=5, P_2=3, P_3=2$, operating on an input sequence $\mathbf{x}=\{x_0, x_2, \ldots, x_{29}\}$. Inputs to each stage are prepended with a trainable start token \textless S\textgreater. The updated patch representations of the input sequence output by the global models are added to the inputs of the next stage. The local model generates individual bytes, and the final outputs are concatenated.}
    \label{fig:mblm-simple}
\end{figure}

\subsection{Stage Models}
All previous work on hierarchical models has been limited to two stages with Transformer decoders as stage block~\cite{megabyte, digitalworldsimulators}. Training a model on this objective requires careful patchification. Specifically, within the same patch, no information to future tokens must be leaked, which MBLM achieves by offsetting the inputs between stages with trainable start tokens. A derived requirement for the models contained within an MBLM is therefore that they are autoregressive. Hierarchical architectures for Transformer decoders have historically aimed to reduce the quadratic cost of self-attention. However, we argue that even models with linear scaling properties like Mamba \cite{mamba} can benefit from compression through patchification. Mamba is a type of state space model (SSM) with a selection mechanism. SSMs describe the evolution of a physical system over time and are parametrized by the system matrix $\mbf{A}\in\R^{N\times N}$, input and output matrices $\mbf{B}\in\R^{N\times1}$ and $\mbf{C}\in\R^{1\times N}$ as well as the step-size $\Delta\in\R$, which defines the resolution of the input. Continuous-time SSMs define a function-to-function map:
\begin{subequations}
    \label{eq:ssm}
    \begin{align}
        \text{State equation}:\;&h'(t)=\mbf{A}h(t)+\mbf{B}x(t)\\
        \text{Output equation}:\;&y(t)=\mbf{C}h(t)+\mbf{D}x(t)
    \end{align}
\end{subequations}
$\mbf{D}x(t)$ can formally be omitted because it constitutes a skip-connection. To discretize the continuous-time system defined in \autoref{eq:ssm}, \citet{mamba} apply the zero-order hold discretization rule, resulting in the discrete parameters $\mbf{\ov{A}}=\text{exp}(\Delta, \mbf{A})$ and $\mbf{\ov{B}}=(\Delta\mbf{A})^{-1}(\text{exp}(\Delta\mbf{A})-\mbf{I})\cdot\Delta\mbf{B}$. 
The discrete SSM can be computed like a recurrent neural network (RNN) that independently maps each channel $D$ of an input $x \in \R^D$ at a time step $t$ to an output $y \in \R^D$ through a higher dimensional latent state $h \in \R^N$~\cite{s4}. Moreover, the recurrence can be unrolled given the initial state $h_0$
% :
% \begin{subequations}
% \begin{align}
% h_0&=\mbf{\ov{B}}x_0 \quad y_0=\mbf{C}\mbf{\ov{B}}x_0\\
% h_1&=\mbf{\ov{AB}}x_0 + \mbf{\ov{B}}x_1 \quad y_1=\mbf{C\ov{AB}}x_0 + \mbf{C\ov{B}}x_1
% \end{align}
% \end{subequations}
and vectorized into the SSM convolutional kernel $\ov{\mbf{K}}$ \cite{s4} for a sequence length $L$:
\begin{subequations}
\begin{align}
    \ov{\mbf{K}}\in\R^L&=(\mbf{C\ov{B}},\mbf{C\ov{AB}},\cdots,\mbf{C}\ov{\mbf{A}}^{L-1}\ov{\mbf{B}})\\
    y&=\ov{\mbf{K}}*x
\end{align}
\end{subequations}
The use of a convolutional kernel for efficiency requires that (1) the model is \emph{linear time-invariant} (LTI), meaning that $\Delta, \mbf{\ov{A}}$ and $\mbf{\ov{B}}$ are fixed for all time steps and (2) $\mbf{A}$ is structured, with the most popular form of structure being diagonal \cite{ssmstructure}. This class of SSM is called \emph{structured} SSM \cite{s4}. In contrast, Mamba is a \emph{selective} SSM that makes the parameters $\Delta, \mbf{B}$ and $\mbf{C}$ functions of the input via linear projections. To compute the time-varying parameters efficiently, the model cannot use a convolution, which assumes a fixed kernel. Instead, it leverages a \emph{parallel associative scan} \cite{nvidiaparallelscan} as part of a hardware-aware algorithm with linear complexity that computes the model during training when the entire sequence is seen at once. During inference, Mamba passes the hidden state through its recurrence mechanism, enabling efficient, RNN-like autoregressive generation with constant time complexity per step. Despite being a recent development, Mamba has surpassed previous state-of-the-art models, including optimized Transformer baselines, on various long-sequence benchmarks, achieving up to 5x higher inference throughput compared to Transformers \cite{mamba}. Mambas have already been scaled to billions of parameters \cite{nvidiaempiricalmambahybrid} and incorporated into hybrid architectures that integrate attention and SSM layers \cite{ssmjamba, glorioso2024zambassm}, or combine Mamba with \emph{mixture-of-experts} approaches \cite{anthony2024blackmamba}. Since its introduction, the original Mamba-1 model has undergone revisions, which resulted in the conceptually similar yet more hardware-efficient Mamba-2 \cite{mamba2}.

\subsection{Datasets \& Evaluation}
\begin{figure}[t!]
    \centering
    \subfigure[Original CLEVR image]{\label{fig:a}\includegraphics[width=0.48\linewidth]{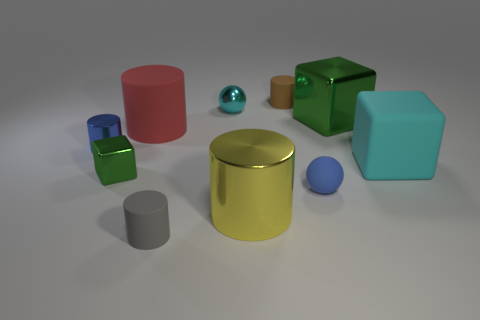}}
    % \label{fig:clevr-img-orig}
    \subfigure[Discretized to 8 colors]{\label{fig:b}\includegraphics[width=0.48\linewidth]{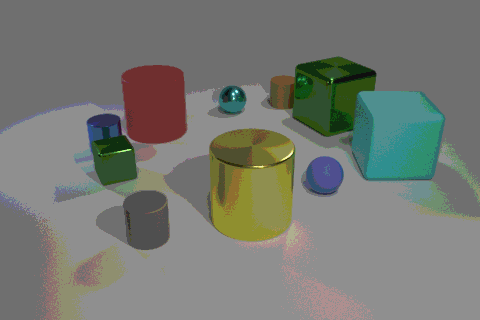}}
    \vspace{-3mm}
    \caption{Each CLEVR image is annotated with multiple questions, e.g.: \textit{What is the shape of the tiny brown thing?} $\rightarrow$ \textit{Cylinder}}
    \label{fig:clevr-img}
\end{figure}

We evaluate the performance of MBLMs in terms of language modeling on the Project Gutenberg (PG19) dataset \cite{pg19}. 
PG19 contains 28,752 English-language books, or 11.6 GB of text, which were published before 1919. 
We select this dataset for comparability to prior art~\cite{megabyte,mambabyte} and because, on average, each book is around 411 KB, which allows long-range language modeling on consecutive bytes in the same document. 
For the multimodal evaluation, we train models on CLEVR \cite{clevr}, a labelled dataset for a visual question answering. CLEVR contains 70,000 synthetically generated RGB images containing 3D shapes, 28 unique answers and roughly 700,000 questions requiring perceptual abilities such as recognizing or counting objects, inferring relationships or making comparisons, with an example provided in \autoref{fig:clevr-img}.
Additional statistics are included in \autoref{sec:app:dataset}.

We use bits-per-byte (BPB) \cite{pile} as the primary evaluation metric for byte-level modeling and report word-level perplexities (PPL) to facilitate comparisons with future work. BPB, related to perplexity, quantifies the average number of bits needed to encode each byte of data and can be seen as a compression measure where a lower value indicates a higher probability of correctly predicting the next byte \cite{scalinglanguagemodelsbpb}. 
\begin{equation}
    \text{BPB}=\log_2(e^{\ell_\text{byte}})=\frac{\ell_\text{byte}}{\ln{2}}
\end{equation}
where $\ell_\text{byte}$ is the observed average negative log-likelihood stemming from a byte vocabulary.
All MBLMs are matched to 360M parameters\footnote{Apart from the 5M 3D MBLM which is a 350M model.} and trained on 8 NVIDIA A100 SXM4 80 GB GPUs in parallel using a custom-built distributed PyTorch trainer. For each experiment, we follow a data-parallel approach and split the training datasets among the GPUs.
Further details on the model and training as well as evaluation metrics are provided in \autoref{sec:app:model-details},~\ref{sec:app:training-recipes} and~\ref{sec:app:evaluation}.
The source code is publicly available for reproduction:~\url{https://github.com/ai4sd/multiscale-byte-lm} and can be installed as \texttt{mblm} directly from PyPI.

\section{Results}
For modeling long byte sequences, we take advantage of MBLMs' ability to combine different models at each stage and combine Transformer decoder with Mamba-2 models in different constellations. All models are referenced by their dimensionality; 1D MBLMs contain only a single stage with either a Transformer decoder or Mamba-2 model. In our implementation, a single-stage hierarchy operates with numerical equivalence to the model when used independently of the hierarchy.

\subsection{Scaling Byte Language Models}
\begin{figure}[h!]
    \centering
    \includegraphics[width=0.9\linewidth]{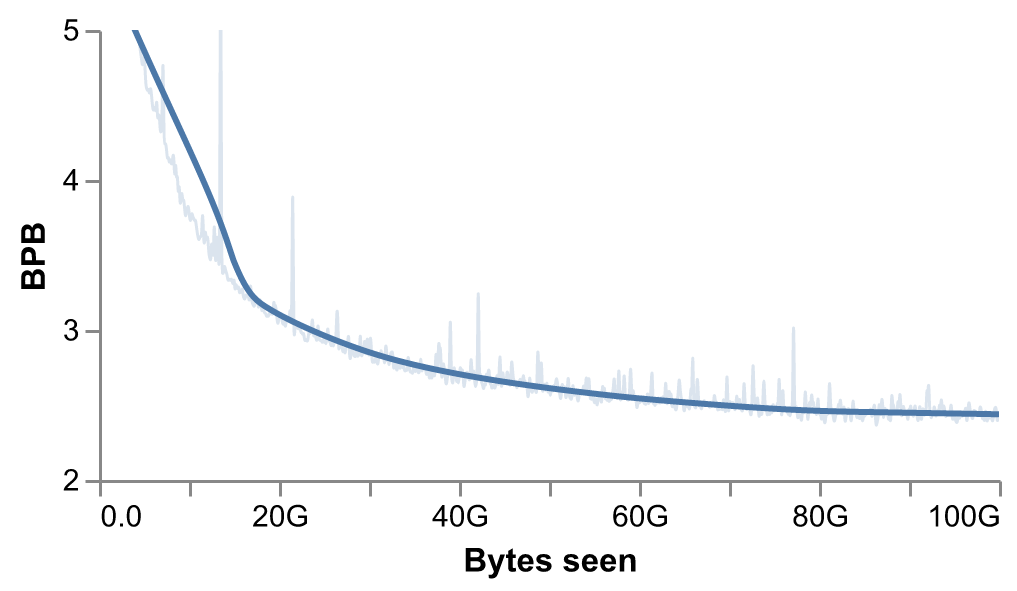}
    \caption{Training loss progression of a 3D MBLM with 350M parameters and a context window of 5 million bytes on a single GPU.}\label{fig:ultrascale}
\end{figure}

As the first three-stage (3D) hierarchical model of its kind, an MBLM comprising a global Mamba followed by two Transformer decoders can process byte sequences of 5 million bytes during training on a single A100 80 GB GPU with standard automatic mixed precision. After just over 15 hours of training this 350M-parameter model processed 100 GB of UTF-8 bytes and achieved 2.448 BPB on the PG19 test set (\autoref{fig:ultrascale}).

By employing a multiscale hierarchy with MBLMs, we target training sequence lengths that ordinary, non-hierarchical (1D) sequence models cannot process without exhausting GPU memory. As shown in \autoref{tab:hierarchical-mem}, the same Transformer decoder backbone scales to twice the sequence length when incorporated into a two- or three-stage MBLM, owing to optimized computational efficiency through input compression. 
\begin{table}[h!]
  \centering
  \small
    \begin{tabular}{lccc}
    \toprule
    MBLM \textbackslash\ Context size & 8192 & 16384 & 32768 \\ 
    \midrule
    1D Transformer & 30.5 & 56.2 & \textit{out of memory} \\ 
    2D Transformer & 19.6 & 35.8 & 68.2 \\
    3D Transformer & \textbf{15.9} & \textbf{28.2} & \textbf{53.0} \\
    \bottomrule
    \end{tabular}
    \caption{Absolute memory usage in GB during training of three 360M parameter Transformer MBLMs on a single NVIDIA A100 80 GB GPU. Hierarchical Transformers scale to 2x the sequence length. All models received batches of 2 sequences of the corresponding length.}
    \label{tab:hierarchical-mem}
\end{table}

Naturally, since MBLMs scale by compressing the input sequence, regular 1D models outperform hierarchical models when the sequence fits into memory. This underscores that hierarchical architectures are specifically designed for extremely long-sequence modeling.
\subsubsection*{Performant Hierarchies}
When configured as a two-stage (2D) hierarchy with two Transformer decoders, an MBLM aligns with the MegaByte architecture \cite{megabyte}. Using the same patch sizes of (8192, 12) for the global and local models, respectively, both hybrid and Mamba-based MBLMs outperform a Transformers-based MegaByte model when trained on 200 GB of PG19 text (\autoref{tab:mblm-vs-mb-2d-100k}). 
\begin{table}[h!]
  \centering
  \small
    \begin{tabular}{llcc}
    \toprule
    Hierarchy & Global \& local model & Test PPL & Test BPB\\ 
    \midrule
    MegaByte & Transformer (2x) & 278.79 & 1.370\\ 
    MBLM & Mamba, Transformer & 163.29 & 1.240\\ 
    MBLM & Mamba, Mamba & \textbf{119.37} &\textbf{1.164 }\\
    \bottomrule
    \end{tabular}
    \caption{Comparison of the MegaByte \cite{megabyte} and MBLM architectures on byte sequences of length 98,304. Hybrid and Mamba-based MBLMs outperform MegaByte on the same amount of data.}
    \label{tab:mblm-vs-mb-2d-100k}
\end{table}

Unlike previous hierarchical architectures, MBLMs can be configured with an unlimited amount of stages and different decoder models at each stage. On context windows exceeding 1 million bytes, hybrid hierarchies again outperform Transformer-based MBLMs (\autoref{tab:mblm-3d-1m}).
\begin{table}[h!]
  \centering
  \small
    \begin{tabular}{lcc}
    \toprule
    3D MBLM configuration & Test PPL & Test BPB\\ 
    \midrule
    Transformer (3x) & 5420.66 & 2.092\\
    Mamba, Transformer (2x) & 5351.71 & \textbf{2.089}\\ 
    \bottomrule
    \end{tabular}
    \caption{After training on 200 GB with a context window of more than 1M bytes (1,048,576), hybrid MBLMs with a first global Mamba perform slightly better than homogeneous Transformer hierarchies.}
    \label{tab:mblm-3d-1m}
\end{table}

To fit the 3D models in \autoref{tab:mblm-3d-1m} on a single GPU, we use a physical batch size of 1. With inner model context sizes of (8,192, 16, 8), the input tensor at stage 3 is given by $\mbf{x}_3\in\R^{131072 \times 8 \times D_3}$. Previous multiscale models like MegaByte \cite{megabyte} advocate for full parallelism at every stage. Yet, full parallelism is often infeasible for extremely long inputs, even on modern GPUs. By enabling the batch chunking feature of MBLMs, the batches for the second and third stages are divided into 10 and 20 chunks, respectively, and intermediate activations for each chunk are recomputed during the backwards pass. This enables each MBLM to train at approximately 75-80\% memory utilization on a single A100 80 GB GPU.

\subsubsection*{Computational Efficiency}
\begin{figure}[h!]
    \centering
    \includegraphics[width=1\linewidth]{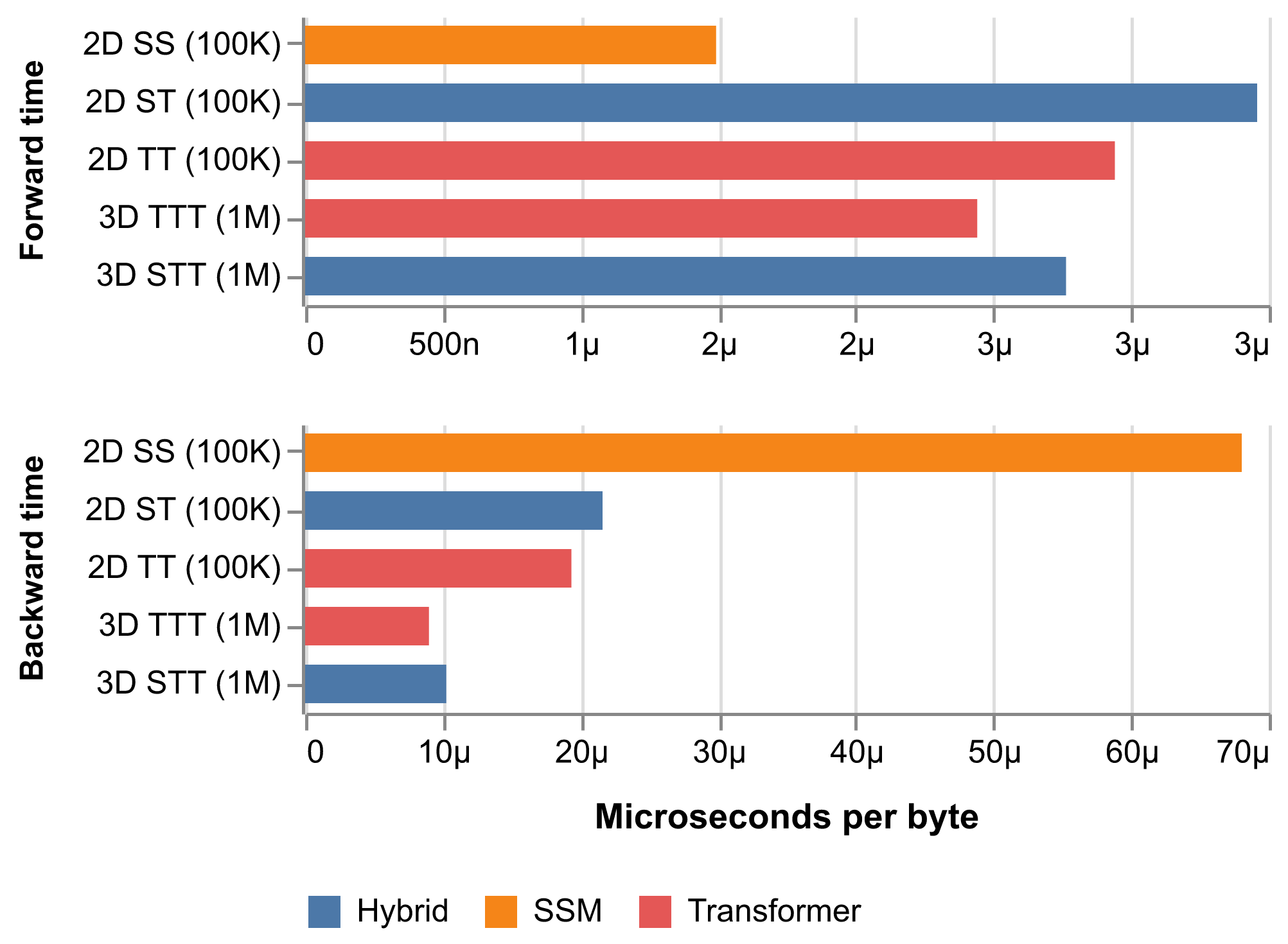}
    \caption{Throughput expressed as time-per-byte for 2D and 3D MBLMs during training. Using Mamba as a local model on short patches results in a 4x slower backwards phase.}\label{fig:fw-bw}
\end{figure}

While above results show that purely Mamba-based MBLMs deliver the best performance, employing Mamba as the local model in a hierarchical configuration is computationally  expensive. Given a 100K byte input sequences, the local SSM inside a 2D MBLM operates on patches of only 8 bytes. Using Mamba on such short sequences results in a 4x longer backwards phase during training compared to an equivalent MBLM with a local Transformer decoder, as shown in~\autoref{fig:fw-bw}.

We attribute this discrepancy to the $\mathcal{O}(BLDN)$\footnote{$B$ represents the batch size, $L$ the sequence length, $D$ the number of input channels and $N$ the SSM state dimension.} time and memory complexity of Mamba's parallel scan. Although linear in $L$, the parallel scan is significantly slower than self-attention, which scales with $\mathcal{O}(L^2D)$ \cite{selfattncomplexity}, for small values of $L$. \citet{mamba} report that Mamba-1's parallel scan is faster than FlashAttention \cite{dao2023flashattention2} for sequence lengths exceeding 2K, emphasizing that Mamba models are specifically designed for efficient modeling of \emph{long} sequences. % Furthermore, it is possible that Mamba's  approach of recomputing the intermediate states during the backwards pass is more noticeable for short sequences.

\subsubsection*{Inference Context Extrapolation}
\begin{figure}[h!]
    \centering
    \includegraphics[width=0.9\linewidth]{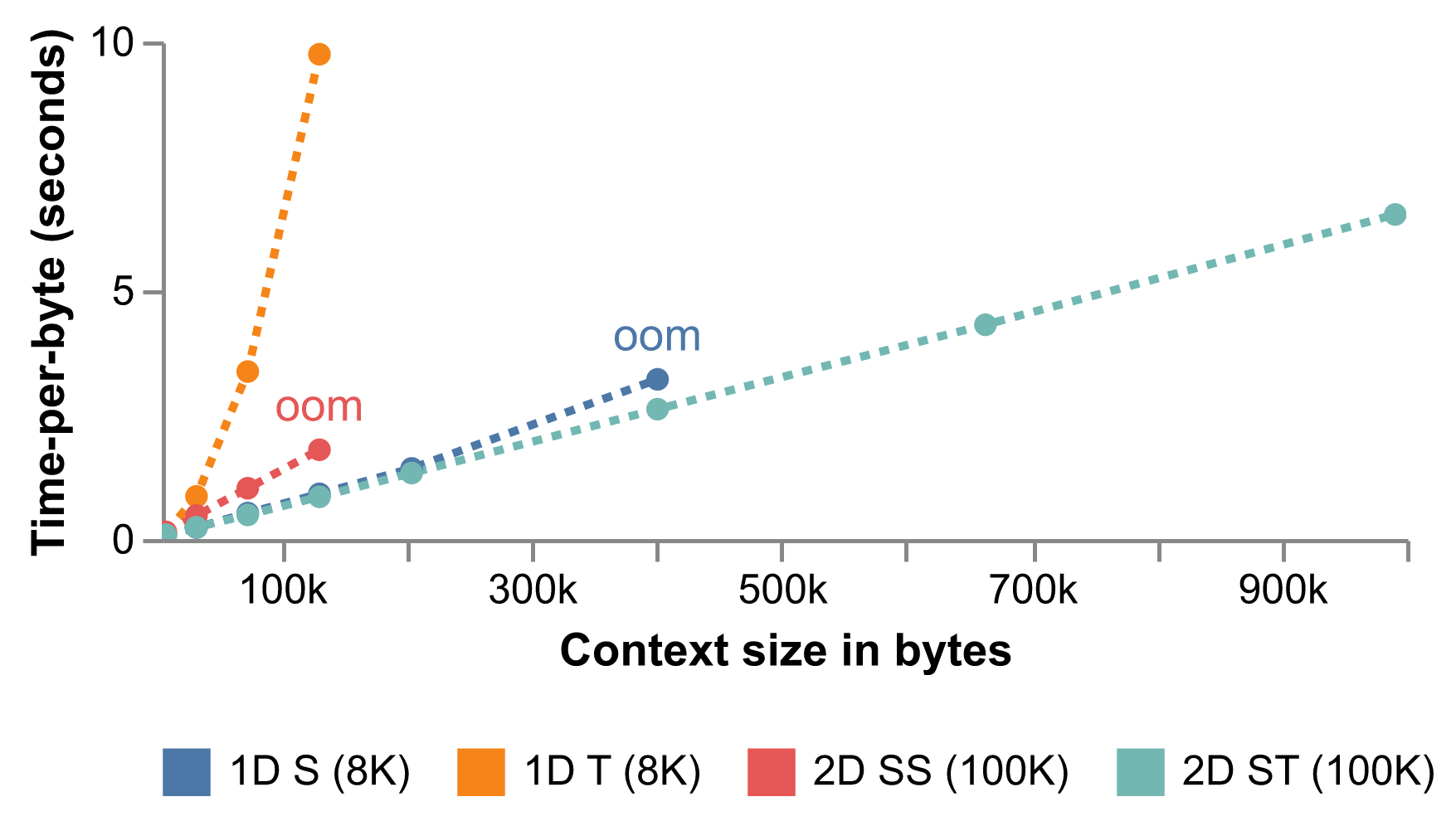}
    \caption{The time it takes to generate a single byte as a function of context size for 1D and 2D MBLMs. Hybrid MBLMs exhibit near-linear generational efficiency.}\label{fig:gen-time}
\end{figure}

To investigate inference throughput and context extrapolation capabilities, we evaluate four different MBLMs on byte input sequences ranging from 8,192 to 991,232 in length $L$. These include two 1D modules trained with an 8K context window and two 2D modules trained with a 100K context window. Since the 1D Transformer uses rotary position embeddings \cite{roformer}, input length is only bound by compute requirements. Efficient inference solutions for 1D models, such as key-value caches for Transformers \cite{ott-etal-2019-fairseq}, have been widely adopted. In its recurrent mode, Mambas can even process each step in constant time by passing the SSM state through the recurrence. However, implementing a dedicated inference pipeline in a hierarchical setting poses significant challenges because patches form a compressed representation of chunks of the input sequence, making it infeasible to cache and reuse previously computed results effectively. As a result, all MBLMs containing a Mamba-2 block still compute a parallel scan over the sequence during inference. While both SSM representations are expected to be numerically equal, this results in longer generation times per token and constrains the model's scalability linearly with respect to the context size. \autoref{fig:gen-time} visualizes the time-per-byte for 1D and 2D MBLMs as a function of context length. 
This result demonstrates that hybrid hierarchies with a global Mamba and local Transformer decoder are able to generate tokens with near-linear efficiently up to a context size of one million bytes.
Instead, generating bytes on extended context windows quickly becomes infeasible for regular Transformers due to their $\mathcal{O}(L^2)$ complexity.
\begin{figure}[h!]
    \centering
    \includegraphics[width=0.9\linewidth]{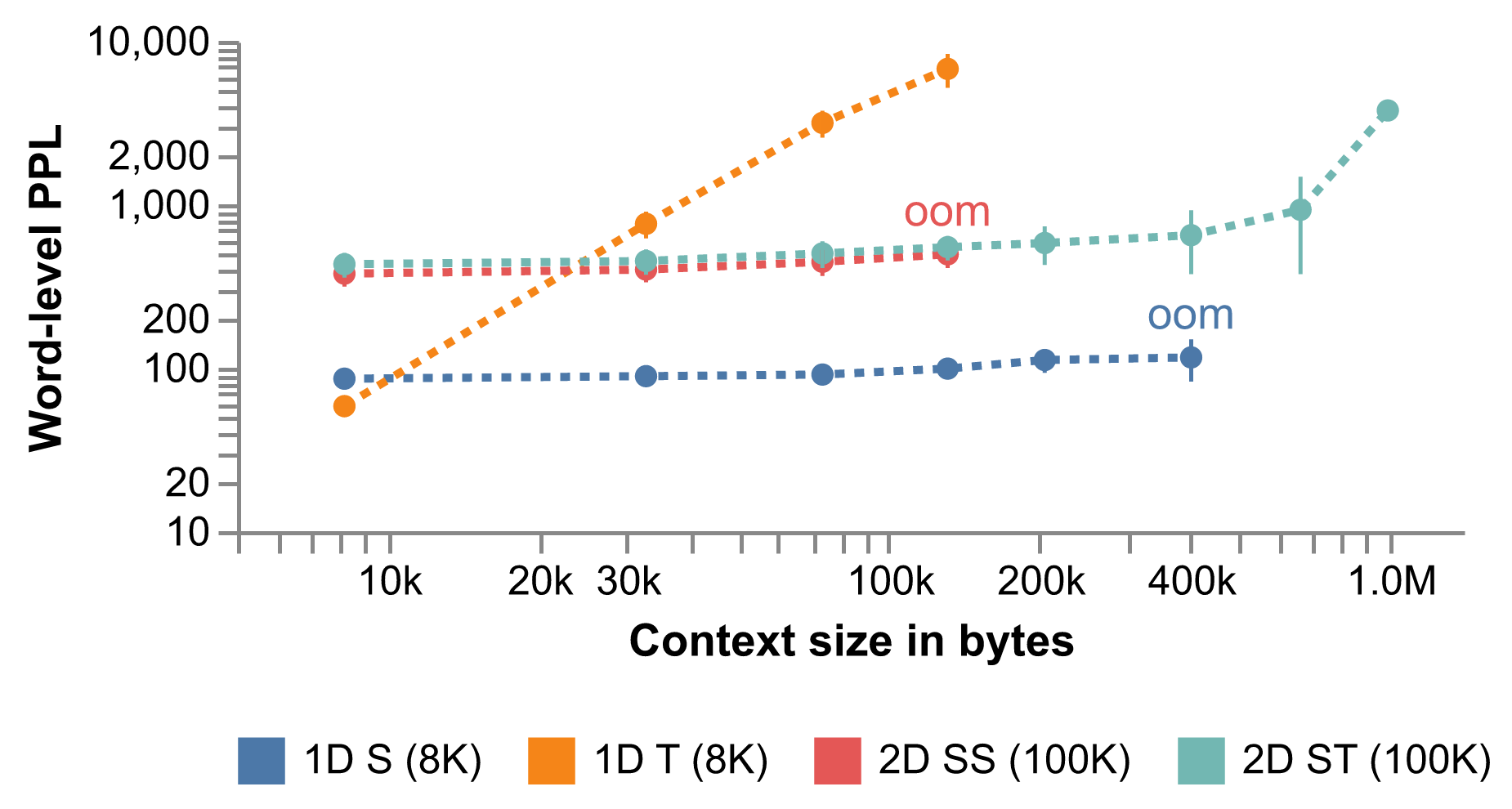}
    \caption{Word-level perplexities as a function of context size for 1D and 2D MBLMs.}\label{fig:gen-ppl}
\end{figure}

On extended context windows, (\autoref{fig:gen-ppl}), our a priori expectation is that both the 1D Mamba and Transformer will perform poorly when evaluated on a context length extended by a factor of 120 \cite{translengthextra, mambaextrapolation}. To our surprise, long context models are not better with more context. This suggests that the models' prediction confidence is largely unaffected by context size, meaning that much of the context is effectively ignored by the models. 
We hypothesized that this is due to the nature of the PG19 datasets and questioned its suitability for assessing large context extrapolation by conducting an ablation study with a Llama 2-7B model~\cite{touvron2023llama2openfoundation}, which has been pre-trained on a context size of 4,096 (subword) tokens and focus on small context sizes up to 8,192 bytes. Details on the conversions are given in \autoref{sec:app:llama-ppl}. 
\begin{figure}[h!]
    \centering
    \includegraphics[width=0.9\linewidth]{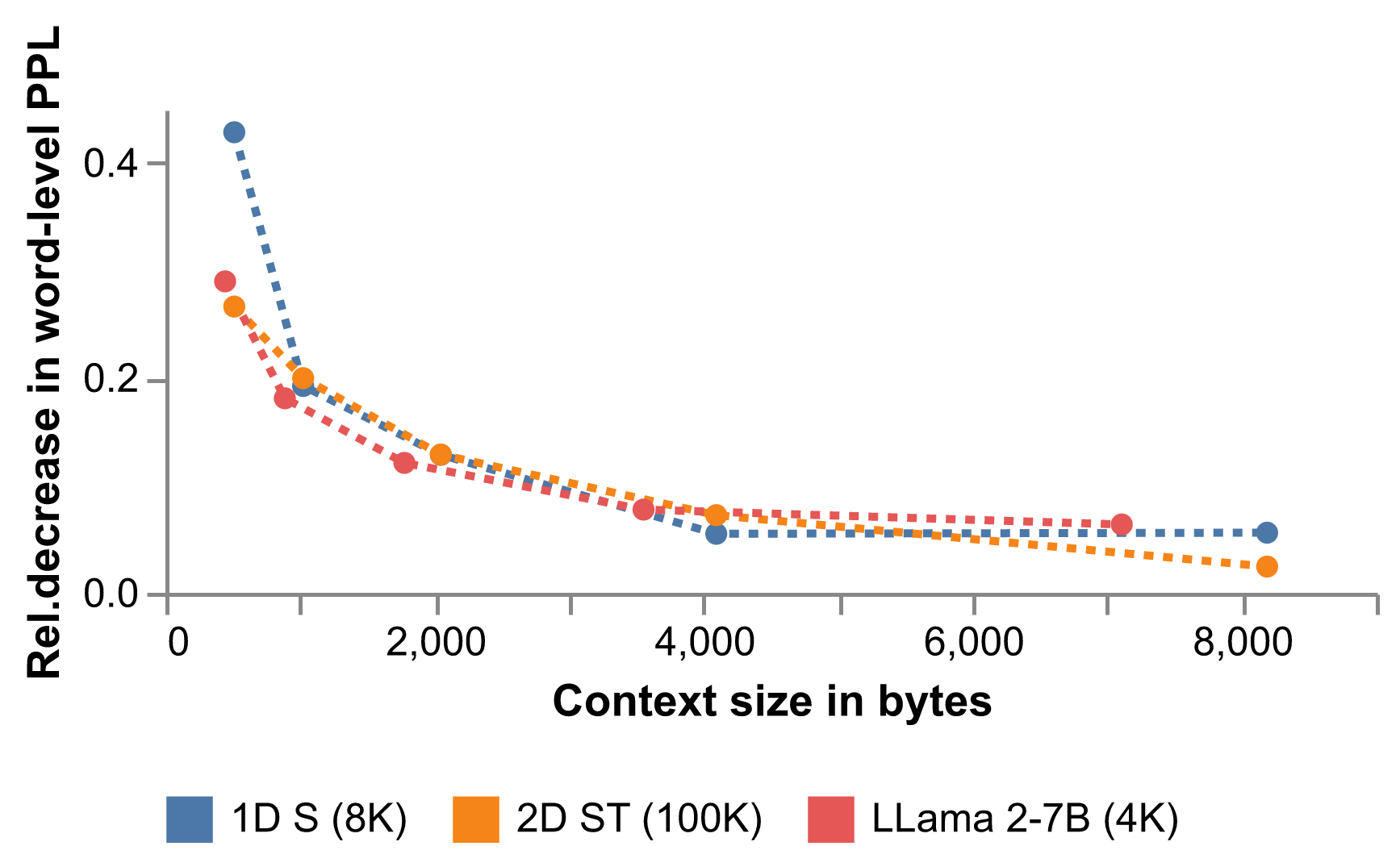}
    \caption{Relative improvement in word-level perplexities for consecutive context lengths for the 1D SSM, 2D SSM-Transformer and Llama baseline}\label{fig:gen-ppl-llama}
\end{figure}

\autoref{fig:gen-ppl-llama} shows the relative improvement in word-level perplexities for consecutive context lengths for the 1D SSM, 2D SSM-Transformer and Llama baseline: On PG19, all models perform strictly better the larger the context size is. However, given a context length of $\geq$ 4,000 bytes, the relative decrease in perplexity diminishes even for a heavily optimized language model such as LLama, indicating that around 4K bytes are likely enough to reasonably predict the next few bytes in a PG19 book.

\subsection{Byte-level Visual Question Answering}
Byte-level models have primarily been applied to text \cite{mambabyte, byt5tokenfree, bytelevelmachinetrans}.
Even though they are ultimately motivated by the generality of bytes as universal data representation, previous research beyond text have focused on single modalities at a time (e.g., audio~\cite{digitalworldsimulators}, images~\cite{hortonbytesareallyouneed} or on modality conversion~\cite{li2019bytes,digitalworldsimulators}.
Suprisingly, no previous work has explored a multimodal task like visual Q\&A with a BLM.
% limited research has explored their application to multimodal tasks.
Our MBLM is naturally well-suited to operate on extremely long bytestreams obtained from multimodal data. A 3D MBLM with a context size of 500K bytes can learn the task of visual question answering on the CLEVR dataset \cite{clevr} given the entire, flattened $480\times320\times3$ RGB image tensor and UTF-8 encoded question (see Appendix~\autoref{fig:app:clevr-qa-bridge}). 
We are the first to benchmark byte-language models on this task, and rather than using an encoder-based approach with average pooling and a classification head for image classification \cite{digitalworldsimulators}, our autoregressive MBLMs employ a language modeling head to predict individual bytes mapped to one of the 28 possible answers in CLEVR. 
While this significantly increases task difficulty -- with a random baseline achieving only $\frac{1}{256} = 0.4\%$ accuracy compared to $\frac{1}{28} = 3.6\%$ -- the language modeling head provides superior generalization capabilities.

The 3D 500K MBLM achieves an accuracy of 44\% on CLEVR's validation set after only 100K samples ($\approx\frac{1}{6}$ of an epoch). 
To mitigate the information overload induced by the images -- over 99.9\% of the input sequences consists of RGB pixel values -- we experiment with different image representations at the byte level. \autoref{tab:clevr-acc} displays the accuracies for 1D Transformer- and Mamba-based MBLMs using different image representations as well as three baselines taken from \citet{clevr}. While the \textit{CNN+LSTM} baseline processes both the image and question and uses an classification head to derive a label, the \textit{LSTM} only looks at word embeddings of the question and \textit{Q-Type} predicts the most frequent answer from the training for the corresponding question type. All our autoregressive MBLMs outperform the LSTM and Q-Type baseline, and, when receiving discretized images, perform comparably to the CNN+LSTM baseline, even in the absence of an image encoder.
% \begin{table}[h!]
%     \small
%     \centering
%     \begin{tabular}{l r r r}
%     \toprule
%     Model & Overall & Exists & Count\\
%     \midrule
%     CNN+LSTM & \textbf{52.3} & 65.2 & \textbf{43.7}\\
%     1D T (DISC) & 52.1 & 69.0 & 39.7\\
%     1D S (DISC) & 51.6 & 68.7 & 38.5\\
%     1D S & 50.3 & 69.7 & 38.0\\
%     1D S (JPEG) & 50.3 & \textbf{72.0} & 39.3\\
%     1D T & 50.0 & 65.3 & 38.3\\
%     1D T (JPEG) & 49.1 & 68.7 & 39.0\\
%     LSTM & 46.8 & 61.1 & 41.7\\
%     Q-Type & 41.8 & 50.2 & 34.6
%     \vspace{7pt}\\
%     \toprule
%     Model & Comp. Int. & Comp. Attr. & Query Attr.\\
%     \midrule
%     CNN+LSTM & 66.0 & \textbf{53.0} & \textbf{49.2}\\
%     1D T (DISC) & 62.2 & 50.8 & 44.6\\
%     1D S (DISC) & 63.4 & 49.9 & 43.4\\
%     1D S & 64.6 & 48.3 & 39.8\\
%     1D S (JPEG) & 64.1 & 51.2 & 36.5\\
%     1D T & 63.6 & 50.6 & 38.3\\
%     1D T (JPEG) & 59.8 & 49.7 & 38.1\\
%     LSTM & \textbf{69.0} & 51.2 & 36.8\\
%     Q-Type & 51.0 & 51.2 & 36.0\\
%     \bottomrule
%     \end{tabular}
%     \caption{Accuracies on the CLEVR validation set by question type. \textit{1D T} and \textit{1D S} models correspond to our Transformer- and Mamba-based MBLMs, respectively. The \textit{Q-Type, LSTM} and \textit{CNN+LSTM} baselines are taken from \citet{clevr}.}
%     \label{tab:clevr-acc}
% \end{table}

\begin{table}[h!]
    \small
    \centering
    \begin{tabular}{l r r r r r | r}
    \toprule
    Model & \textbf{E} & \textbf{C} & \textbf{CI} & \textbf{CA} & \textbf{QA} & All \\
    \midrule
    Q-Type & 50.2 & 34.6 & 51.0 & 51.2 & 36.0 & 41.8 \\
    LSTM & 61.1 & 41.7 & 69.0 & 51.2 & 36.8 & 46.8 \\
    CNN+LSTM & 65.2 & \textbf{43.7} & \textbf{66.0} & \textbf{53.0} & \textbf{49.2} & \textbf{52.3}\\
    \midrule
    1D T (DISC) & 69.0 & 39.7 & 62.2 & 50.8 & 44.6 & 52.1 \\
    1D S (DISC) & 68.7 & 38.5 & 63.4 & 49.9 & 43.4 & 51.6 \\
    1D S & 69.7 & 38.0 & 64.6 & 48.3 & 39.8 & 50.3 \\
    1D S (JPEG) & \textbf{72.0} & 39.3 & 64.1 & 51.2 & 36.5 & 50.3 \\
    1D T & 65.3 & 38.3 & 63.6 & 50.6 & 38.3 & 50.0 \\
    1D T (JPEG) & 68.7 & 39.0 & 59.8 & 49.7 & 38.1 & 49.1 \\
    \bottomrule
    \end{tabular}
    \caption{Accuracies on the CLEVR validation set by question type. Columns: \textbf{E} = Exists, \textbf{C} = Count, \textbf{CI} = Compare Integer, \textbf{CA} = Compare Attribute, \textbf{QA} = Query Attribute, and the final column is the overall accuracy. The \textit{1D T} and \textit{1D S} models correspond to our Transformer- and Mamba-based MBLMs, respectively. The \textit{Q-Type, LSTM}, and \textit{CNN+LSTM} baselines are taken from \citet{clevr}.}
    \label{tab:clevr-acc}
\end{table}
Interestingly, on the \textit{\textbf{E}xists} question type, our MBLMs consistently outperform all baselines.
We hypothesize that this is because the other question types (\textit{\textbf{C}ounting}, \textit{\textbf{C}omparing} \textit{\textbf{I}ntegers} or \textit{\textbf{C}omparing} \textit{\textbf{A}ttributes}) require integraton of spatial information across multiple locations, whereas \textit{\textbf{E}xists} questions can be answered without analyzing spatial relationships.
Note that MBLMs are (1) modality-agnostic, (2) do not possess encoder blocks and (3) entirely lack spatial information (not even patch embeddings or 2D positional encodings). In addition, the row-wise raster scan used to flatten images, combined with the unidirectional modeling approach, can make it impossible to associate information from specific regions of the original image, depending on the arrangement of the scene and the nature of the question.
Thus spatial reasoning becomes challenging, yet MBLMs perform better than standard models at finding singular pieces of information as in the \textit{\textbf{E}xists} task. 

Using the inherent knowledge transfer capabilities of byte-level models, we fine-tuned the 1D MBLMs from ~\autoref{tab:clevr-acc} on CLEVR data using MBLMs pre-trained on UTF-8 bytes from the PG19 dataset. Contrary to \citet{digitalworldsimulators}, who reported negative transfer effects from UTF-8 text to vision tasks, our results demonstrate that pre-training on text bytes positively impacts mixed-modality VQA performance (see Appendix \autoref{fig:app:ft-vs-nft}).

\subsubsection*{Byte-level image representations}
All 1D MBLMs predict a byte label based on a flat bytestream of length 8,192 that contains an image representation and the UTF-8 encoded question. Some models, denoted with the \texttt{DISC} suffix in \autoref{tab:clevr-acc}, predict the answer from CLEVR images with reduced color depth. Instead of the original 8-bit color depth (256 channel values), images were discretized to 3-bit, resulting in 8 unique color values per image. An example of this preprocessing step is given in \autoref{fig:clevr-img}\textbf{b}. Discretization reduces information overload while preserving most of the visual features, resulting in performance improvements of up to 5\% per question type.\\
Results from \citet{hortonbytesareallyouneed} suggest that working directly with file bytes offers a performant, preprocessing-free modeling approach. Following this approach, we train \texttt{JPEG} variants for the 1D Mamba and Transformer MBLMs. These models learn directly from on a compressed JPEG bytestream obtained from converting the RGB tensors with a quality factor of 12. While learning directly from file bytes has no negative impact across all VQA tasks, we find that for the \textit{\textbf{E}xists} and \textit{\textbf{C}ompare \textbf{I}nteger} tasks, this method even improves accuracies by almost 7\%. The performance increase likely results from the nature of JPEG compression of eliminating high-frequency details while preserving low-frequency features, such as shapes, colors, and overall structure. This makes it easier for models to detect object presence and count discrete entities.

\section{Discussion}
In this work, we introduced the Multiscale Byte Language Model (MBLM), a hierarchical, model-agnostic architecture capable of scaling to byte sequences as long as 5 million bytes on a single GPU. The MBLM hierarchy operates in stages, with independent autoregressive models at each stage. Byte sequences are divided into patches, embedded, and refined as they pass through the hierarchy, culminating in a local model that autoregressively predicts bytes within each patch. This approach enables efficient processing of very long byte sequences through compression.

Our language modeling experiments demonstrated that MBLMs can handle unprecedented sequence lengths. While Mamba-based hierarchies performed best, hybrid models combining Mamba for global stages and Transformer decoders for local stages achieved an optimal balance between performance and computational efficiency. Hybrid models also converged faster and exhibit near-linear generational efficiency during inference. The novel evaluation of byte-language models on the task of visual question answering revealed that autoregressive models can perform competitively to CNN baselines, even with a language modeling head and when learning from a mixed-modality byte stream.

We recommend extending MBLM evaluations to tasks requiring long contexts, such as multimodal document summarization or needle in a haystack tasks and investigating their performance when scaled to billions of parameters. The MBLM architecture, available on GitHub and as a PyPi package, provides a modular and flexible framework for further development. Its scaling capabilities can be enhanced through features like tensor parallelism or model sharding, which can seamlessly integrate into the hierarchy regardless of the stage models used. With the right technical extensions, we believe MBLMs are well-suited to process sequences spanning tens of millions of bytes. These opportunities position MBLMs as a strong foundation for tackling million-scale byte sequence modeling and driving future innovations in hierarchical architectures.

\section*{Impact Statement}
This paper presents work whose goal is to advance the field of Machine Learning.
Our work is particularly focused on improving context window length in language models which may allow future algorithms to sift through larger amounts of data in one shot.  
There are many potential societal consequences of such work, none which we feel must be specifically highlighted here.

\bibliography{example_paper}
\bibliographystyle{icml2025}

%%%%%%%%%%%%%%%%%%%%%%%%%%%%%%%%%%%%%%%%%%%%%%%%%%%%%%%%%%%%%%%%%%%%%%%%%%%%%%%
%%%%%%%%%%%%%%%%%%%%%%%%%%%%%%%%%%%%%%%%%%%%%%%%%%%%%%%%%%%%%%%%%%%%%%%%%%%%%%%
% APPENDIX
%%%%%%%%%%%%%%%%%%%%%%%%%%%%%%%%%%%%%%%%%%%%%%%%%%%%%%%%%%%%%%%%%%%%%%%%%%%%%%%
%%%%%%%%%%%%%%%%%%%%%%%%%%%%%%%%%%%%%%%%%%%%%%%%%%%%%%%%%%%%%%%%%%%%%%%%%%%%%%%
\newpage
\appendix
\onecolumn

\renewcommand{\thefigure}{A\arabic{figure}}
\renewcommand{\thealgorithm}{A\arabic{algorithm}}
\renewcommand{\thetable}{A\arabic{table}}
\setcounter{figure}{0} % Reset figure counter
\setcounter{table}{0} % Reset table counter

\section{Dataset Details}\label{sec:app:dataset}
\begin{figure}[h!]
    \centering
    \includegraphics[width=0.75\textwidth]{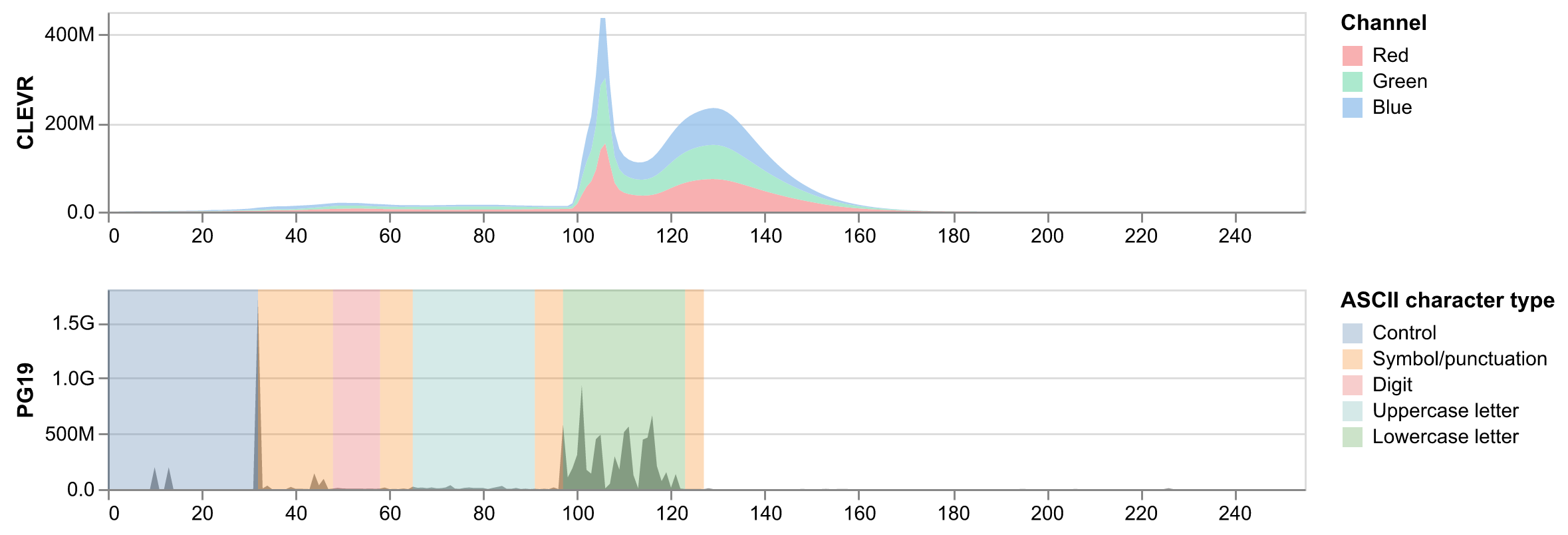}
    \caption{The RGB histogram of images in the CLEVR dataset in comparison to the distribution of UTF-8 bytes contained in the PG19 dataset based on 10 GB of randomly sampled data.}\label{fig:ds-byte-dist}
\end{figure}

For all unimodal experiments, we use a byte vocabulary of $255+1$ tokens, with token ID 257 designated as the \texttt{<pad>} token to enable sequence padding within minibatches. This method aligns with byte-level models like bGPT \cite{digitalworldsimulators}, which employ an \texttt{<eop>} (end-of-patch) token with ID 257 to pad patches. While our \texttt{<pad>} token is not used during training on PG19, it is required during inference to pad patches for prompts that are shorter than the context size. The individual books in PG19, stored in \texttt{.txt} format, are read as bytes from disk and combined into a single \texttt{bytearray} data structure without textual preprocessing. While most PG19 books are within the ASCII character set, some contain Unicode characters outside the ASCII range and are thus encoded in UTF-8. From this byte sequence, we sample subsequences for training based on the context size of the corresponding model. While our data ingestion process is simple and unbiased, the lack of language-specific preprocessing introduces noisy input data, including ASCII control characters like \texttt{NUL} and \texttt{CR} (carriage return), which are usually absent from subword-based vocabularies. A significant portion of the data comprises space characters and newlines. \autoref{tab:pg19} contains statistics for PG19, which we use to derive word-level perplexities. \autoref{fig:ds-byte-dist} compares the distribution of bytes according to 10 GB of randomly sampled data in CLEVR and PG19.
\begin{table}[h!]
  \centering\small
    \begin{tabular}{lccc}
    \toprule
    \centering & $L_B$ & $L_W$ & $L_B / L_W$\\ 
    \midrule
    Train & 11,678,184,667 & 1,966,200,384  & 5.9395\\ 
    Validation & 17,733,002 & 3,007,061  & 5.8971\\ 
    Test & 41,289,101 & 6,966,499 & 5.9268\\
    \bottomrule
    \end{tabular}
    \caption{PG19 \cite{pg19} dataset statistics. $L_B$ is the number of UTF-8 encoded bytes, $L_W$ the number of space-separated words. To count the words, we read all books into a single Unicode string and then split at all common whitespace characters (\symbol{92}n, \symbol{92}r, \symbol{92}t, \symbol{92}f) using Python's \texttt{str.split}.}
    \label{tab:pg19}
\end{table}

\section{Model Details}\label{sec:app:model-details}
All our models pre-trained on the PG19 dataset are matched to 360 million parameters, which is achieved simply by varying the number of layers for each model in the hierarchy. Model names are abbreviated; \textbf{S} stands for Mamba-2 and \textbf{T} for Transformers.

\subsubsection*{Common model configuration}
We keep the model-specific configuration constant: For \textbf{Mamba-2} models, we use a model dimension of $1024$, an SSM state expansion factor $128$, a local convolution width of $4$, a block expansion factor of $2$ and $64$ as the head dimension. Mamba-2 models operate without positional embeddings. \textbf{Transformer} models use a model dimension of $1024$, $16$ attention heads of dimension $64$ and a feed-forward expansion factor of $2$. The attention layers employ rotary positional embeddings (RoPE) \cite{roformer}. Positional encodings are only employed in 2D or 3D multiscale hierarchies for Transformer decoder models. This ensures that all models can be used for context extrapolation experiments. An exception to the default configuration is the 5 million context size experiment, which uses hidden dimensions of size $256$. Throughout all experiments, we used the same context/patch sizes for hierarchical constellations, which are summarized in Table \ref{tab:app:ctx-sizes}.
\begin{table}[h!]
  \centering\small
    \begin{tabular}{l | c c c}
     Context size \textbackslash\ Hierarchy & 1D & 2D & 3D \\ 
    \midrule\midrule
    8192 & 8192 & 1024, 8 & 256, 8, 4\\
    16384 & 16384 & 2048, 8 & 512, 8, 4\\
    32768 & 32768 & 4096, 8 & 1024, 8, 4\\
    98304 & - & 8192, 12 & - \\
    1048576 & - & - & 8192, 16, 8 \\
    5000000$^*$ & - & - & 1000, 200, 25 \\
    \end{tabular}
    \caption{Context/patch sizes across all experiments, denoted from global to local. The MBLM with a 5M context size (denoted with $^*$) uses different model configuration than others, as noted above.}
    \label{tab:app:ctx-sizes}
\end{table}

\subsubsection*{Unimodal experiments}
Table \ref{tab:app:model-layers} summarizes the number of layers for each of the models pre-trained on PG19 \cite{pg19}. We train models with context sizes larger than 98,304 on 200 billion bytes and all others on 30 billion bytes form PG19.

\begin{table}[h!]
  \centering\small
    \begin{tabular}{l | r r r r r r r}
    Model \textbackslash\ Input length & 8,192 & 16,384 & 32,768 & 98,304 & 524,288 & 1,048,576 & 5,000,000 \\ 
    \midrule
    \midrule
    \text{1D T} & 42 & 41 & 39 & - & - & - & -\\
    \text{1D S} & 54 & - & - & - & - & - & -\\
    \text{2D SS} & 24, 28 & - & - & 27, 24 & - & - & -\\
    \text{2D ST} & 25, 21 & - & - & 24, 21 & - & - & -\\
    \text{2D TS} & 25, 20 & - & - & - & - & - & -\\
    \text{2D TT} & 22, 19 & 22, 19 & 22, 19 & 21, 18 & - & - & -\\
    \text{3D STT} & - & - & - & - & 14, 12, 9 & 9, 8, 8 & 1, 1, 1\\
    \text{3D TTT} & 15, 12, 10 & 15, 12, 10 & 15, 12, 10 & - & - & 8, 7, 7 & -\\
    \end{tabular}
    \caption{The number of layers, denoted from global to local, for the 360 million parameter models.}
    \label{tab:app:model-layers}
\end{table}
We do not use MBLMs' gradient checkpointing for 8K models. For all large 100K and 1M models, we use the following number of chunks:
\begin{itemize}
    \item 2D-100K: 2 (stage 2)
    \item 3D-1M: 10 (stage 2), 20 (stage 3)
\end{itemize}

\subsubsection*{Multimodal Experiments}
The multimodal MBLMs in the context of the evaluation on CLEVR are fine-tuned from the corresponding best-performing PG19 MBLMs. However, we use slightly different fine-tuning recipes across all multimodal experiments, as listed in \autoref{sec:app:training-recipes}. 
All 1D MBLMs from the CLEVR section are trained for 3 epochs on shuffled training data, amounting to approximately 2.1 million samples. 
Notably, related work typically trains VQA models on CLEVR for significantly more than 50 epochs \cite{multimodalreason, yi2018neural}.
%wang2021interpretable
The best-performing model for each configuration is selected based on frequent validation set evaluations. Interestingly, all models exhibit peak performance after processing around 450,000 samples (64\% of an epoch), with performance gradually declining thereafter. For each of the 13 VQA question types, we pick 300 random samples to evaluate model accuracies.

\section{Training Recipes}\label{sec:app:training-recipes}
We use slightly different hyperparameters for the unimodal and mulitmodal models, which are listed in Table \ref{tab:app:params}. Prior to our experiments, we validated a few hyperparameters suggested by prior art to train hierarchical models and SSMs respectively:
\begin{description}
    \item[Learning rate] Unlike Megabyte \cite{megabyte}, we find that using a peak learning rate of $1\mathrm{e}{-3}$ results in the best performance on PG19 among the tested 1D and 2D models and other learning rates $4\mathrm{e}{-4}$, $8\mathrm{e}{-4}$
    \item[Positional encodings for Mamba] In preliminary experiments, we test the addition of fixed positional embeddings to the input sequence and find that the SSM performs best without any positional information, regardless of the position in the hierarchy.
\end{description}

\begin{table}[h!]
  \centering\small
    \begin{tabular}{l|cc}
    Parameter & Unimodal (pre-training) & Multimodal (fine-tuning)\\ 
    \midrule
    \midrule
    Learning rate & 0.001 & 0.0001\\ 
    Gradient step & 48 & 84\\ 
    Gradient clipping & 1 & 1\\
    Attention/SSM dropout & 0 & 0.1\\
    \end{tabular}
    \caption{Hyperparameters for the unimodal and multimodal experiments.}
    \label{tab:app:params}
\end{table}
Similar to \citet{megabyte} and \citet{mambabyte}, we use the AdamW optimizer with $\beta = (0.9, 0.95)$ with a linear warmup of 10\% of the total gradient steps followed by cosine annealing. While the physical batch sizes used vary between experiments and are set to maximize GPU efficiency, we use gradient accumulation to arrive at the same gradient step (see Table \ref{tab:app:params}). We keep all models in full \verb|float32| precision and use PyTorch's Automatic Mixed Precision package to enable \verb|float16| precision for the backward passes and the integration of FlashAttention \cite{dao2023flashattention2}. Our Mamba-2 models are built using the \verb|mamba-ssm|\footnote{\href{https://github.com/state-spaces/mamba}{https://github.com/state-spaces/mamba}} (version 2.2.2) and \verb|causal-conv1d|\footnote{\href{https://github.com/Dao-AILab/causal-conv1d}{https://github.com/Dao-AILab/causal-conv1d}} (version 1.4.0) packages. The Transformer models are based on \verb|megabyte-pytorch|\footnote{\href{https://github.com/lucidrains/MEGABYTE-pytorch}{https://github.com/lucidrains/MEGABYTE-pytorch}} (version 0.3.5), which also served as a baseline implementation for MBLM. Any parameter we did not explicitly mention above would use the default value in the corresponding package versions above. We use PyTorch 2.4.1 and train all models on 8 NVIDIA A100 SXM4 80GB GPUs in parallel using a custom-built distributed trainer. For each experiment, we follow a data-parallel approach and split the training datasets among the GPUs.

\section{Evaluation}\label{sec:app:evaluation}
Given the average negative log-likelihood $\ell_\text{subword}$, \citet{pile} define bits-per-byte as:
\begin{equation}
    \text{BPB}=\frac{L_S}{L_B}\log_2(e^{\ell_\text{subword}})
\end{equation}
where $L_S$ and $L_B$ is the length of the dataset in subwords/tokens and length of the dataset in bytes, respectively. If we solely model on bytes, i.e., $L_S = L_B$, this definition can be simplified to:
\begin{equation}
      \text{BPB}=\ell_\text{byte}\log_2 e=\frac{\ell_\text{byte}}{\ln{2}}
\end{equation}
In order to translate between the metrics, we can derive \emph{word-level} perplexities \cite{mambabyte}, which are often used in language modeling. Word-level perplexities are more interpretable and better aligned with human understanding as they measure uncertainty at the level of entire words rather than subwords or bytes. They also facilitate fairer comparisons between models with different tokenization schemes by reducing (though not eliminating) biases introduced by tokenizer differences through normalization. With $L_W$ denoting the number of space-separated words in a corpus, $\text{PPL}_\text{word}$ can be derived from either $\ell_\text{byte}$ or $\ell_\text{subword}$:
\begin{align}\label{eq:ppl}
  &\text{PPL}_\text{word}=\exp{\left(\frac{L_B}{L_W}\ell_\text{byte}\right)} & \text{PPL}_\text{word}=\exp{\left(\frac{L_S}{L_W}\ell_\text{subword}\right)}&
\end{align}
$L_S$, $L_B$ and $L_W$ for PG19 are summarized in \autoref{tab:pg19}. In practice, both PPL and BPB can be understood as scaled variants of the cross-entropy between two distributions. Importantly, minimizing cross-entropy will result in smaller absolute values for both PPL and BPB, which all indicate a more performant model.

\section{Llama-7B Word-Level Perplexities}\label{sec:app:llama-ppl}
We calculate perplexity on subword context sizes varying from 64 to 8,192 on a quantized Llama-2-7B model\footnote{\href{https://huggingface.co/TheBloke/Llama-2-7B-GGUF/blob/main/llama-2-7b.Q5_K_S.gguf}{https://huggingface.co/TheBloke/Llama-2-7B-GGUF/blob/main/llama-2-7b.Q5\_K\_S.gguf}} \cite{touvron2023llama2openfoundation} using the \texttt{llama.cpp} project\footnote{\href{https://github.com/ggerganov/llama.cpp}{https://github.com/ggerganov/llama.cpp}}. We recall that there are a total of $L_W=3,007,061$ whitespace-separated words in the PG19 validation set. Tokenizing the validation set with the SentencePiece-based Llama tokenizer results in $L_S=5,106,780$ subwords. Using the \texttt{llama.cpp} CLI does not give us the direct negative log-likelihood, $\ell_{subword}$, so we have to convert the obtained subword-level \emph{perplexity} values to word-level perplexities by continuing from \autoref{eq:ppl}. Since $\ell_{subword}=\ln(\text{PPL}_\text{subword})$, using basic logarithm rules, we derive:
\begin{equation}
  \text{PPL}_\text{word}=e^{\left(\frac{L_S}{L_W}\ell_{subword}\right)}=e^{(\ln{\text{PPL}_\text{subword}})\frac{L_S}{L_W}}=\text{PPL}_\text{subword}^\frac{L_S}{L_W}
\end{equation}
From the above numbers, $\frac{L_S}{L_W}\approx1.6982$, meaning that a single word in PG19's validation set word consists of approximately 1.7 Llama-subwords. For comparison, \citet{mambabyte} report $\frac{L_S}{L_W}=1.45$ when fitting a SentencePiece tokenizer \cite{kudo-richardson-2018-sentencepiece} on PG19's validation set. We also note that there are $\approx3.4724$ bytes per subword, which we use to convert between subword- and byte-level context lengths.

\section{Additional Figures}
\begin{figure}[!htb]
    \centering
    \includegraphics[width=0.8\linewidth]{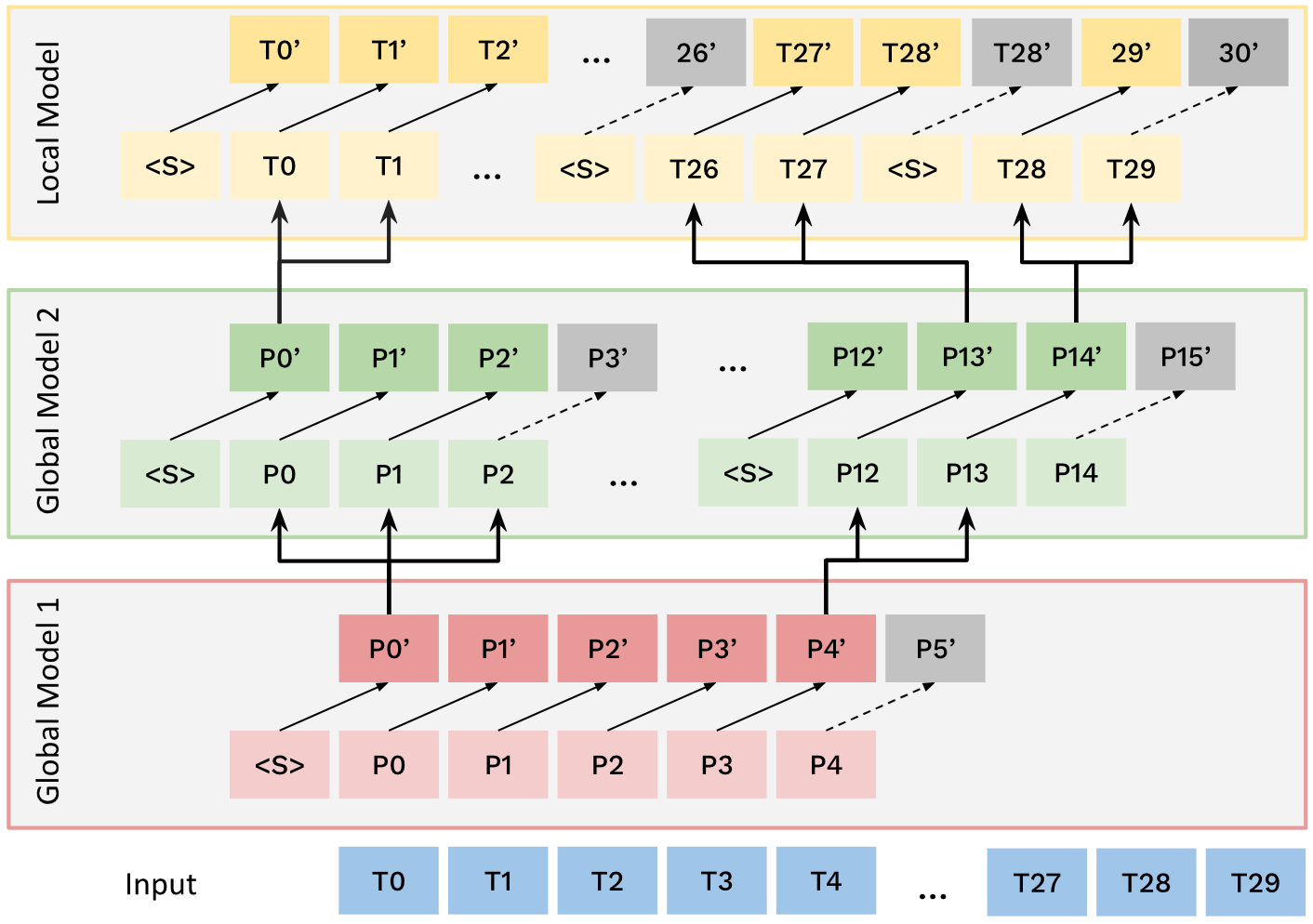}
    \caption{
   A 3D MBLM module with two global and one local decoder models and corresponding patch sizes $P_1=5, P_2=3, P_3=2$, operating on an input sequence $\mathbf{x}=\{x_0, x_2, \ldots, x_{29}\}$. Inputs to each stage are prepended with a trainable start token \textless S\textgreater. The updated patch representations of the input sequence output by the global models are added to the inputs of the next stage. The local model generates individual bytes, and the final outputs are concatenated.}
    \label{fig:mblm-simple}
\end{figure}
\begin{figure}[h!]
    \centering
    \includegraphics[width=0.5\linewidth]{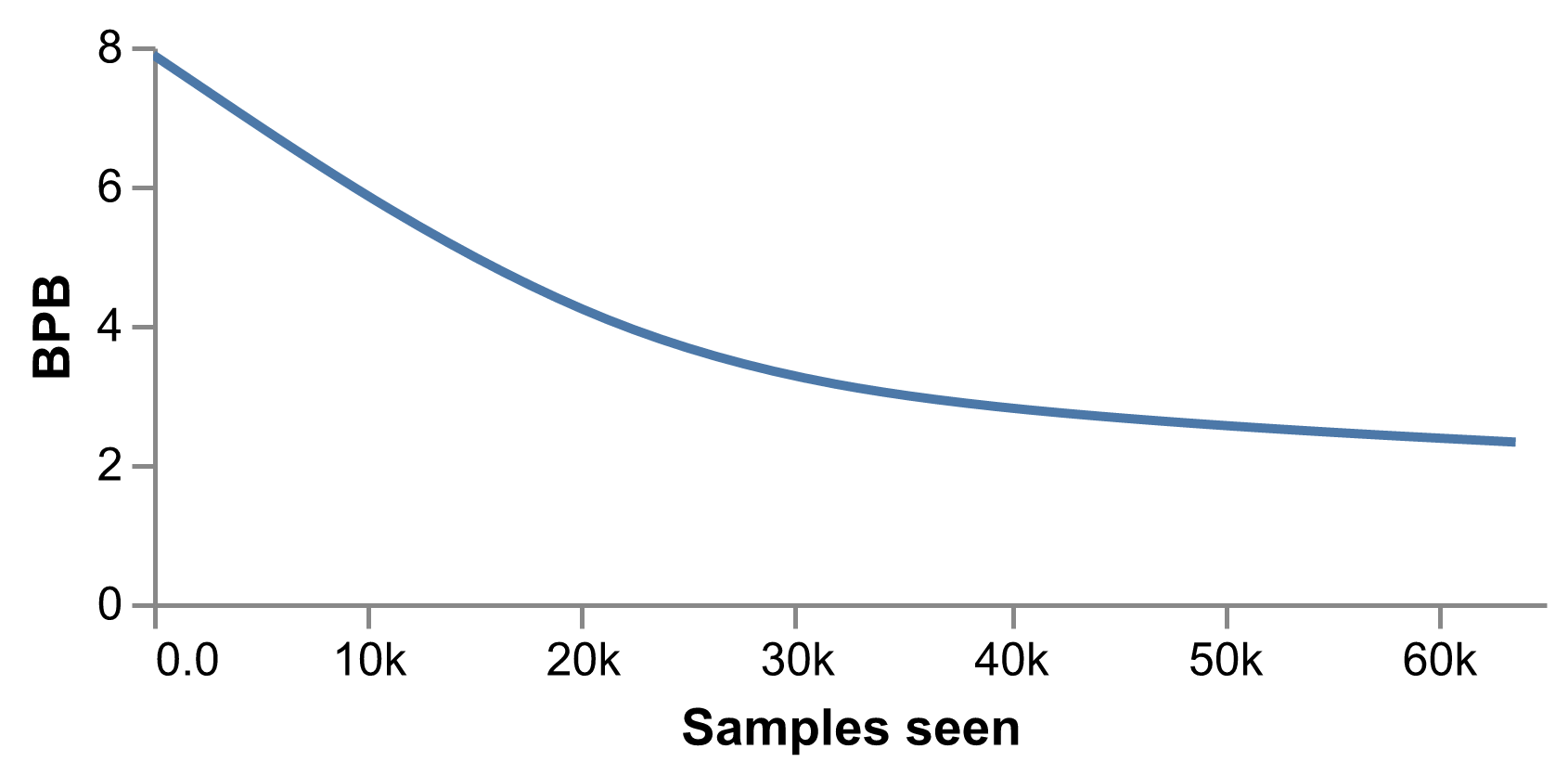}
    \caption{A 3D MBLM on can learn visual question answering directly from unprocessed, flattened RGB images.   }\label{fig:app:clevr-qa-bridge}
\end{figure}
\begin{figure}[h!]
    \centering
    \includegraphics[width=0.5\textwidth]{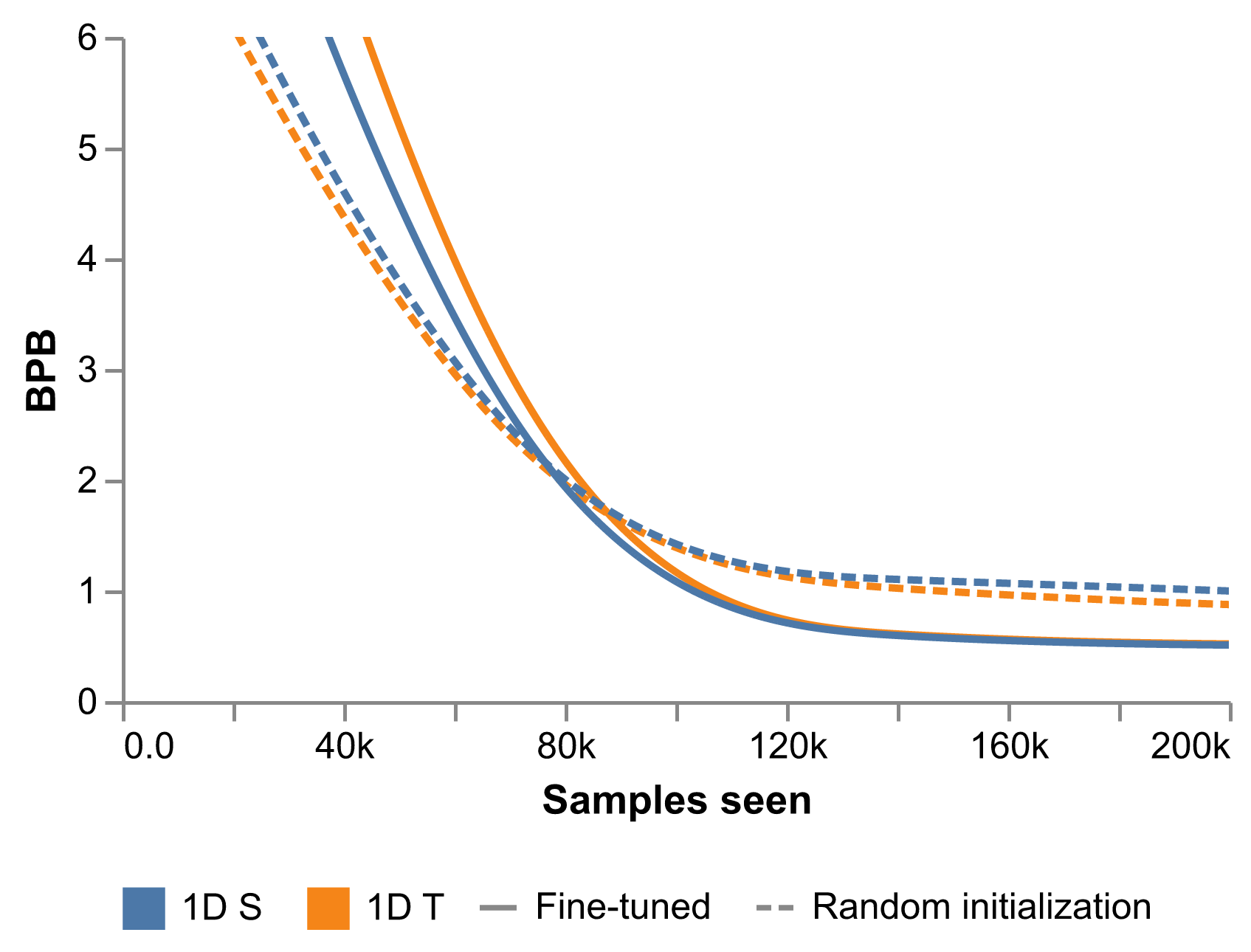}
    \caption{Fine-tuning 1D MBLMs on mixed-modality VQA data, starting from models pre-trained on UTF-8 bytes from the PG19 dataset, reveals positive knowledge transfer effects compared to models initialized with random weights.}\label{fig:app:ft-vs-nft}
\end{figure}

\begin{figure}[h!]
    \centering
    \includegraphics[width=0.85\textwidth]{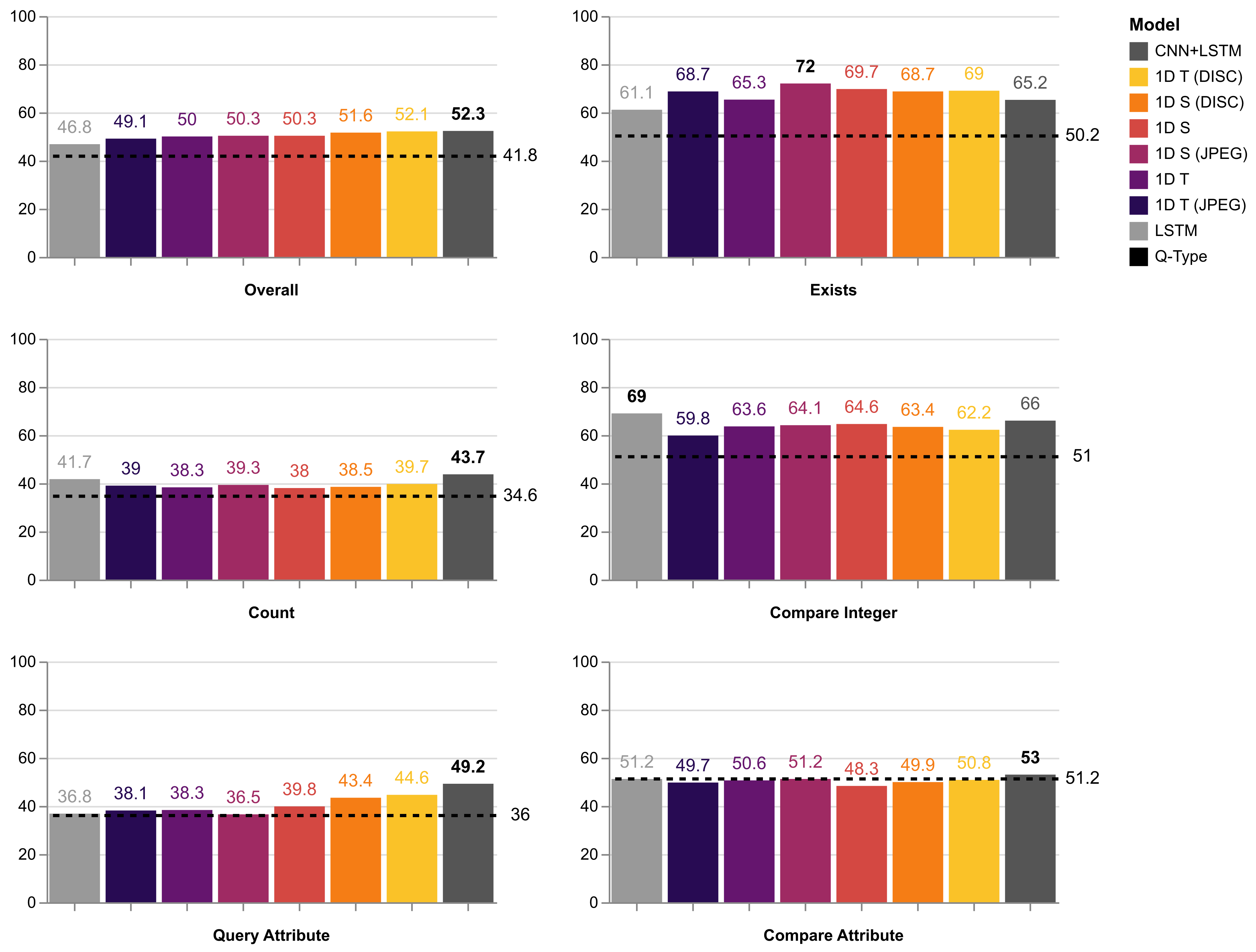}
    \caption{Accuracies on the CLEVR validation set by question type. \textit{1D T} and \textit{1D S} models correspond to our Transformer- and Mamba-based MBLMs, respectively. The \textit{Q-Type, LSTM} and \textit{CNN+LSTM} baselines are taken from \citet{clevr}.}\label{fig:app:clevr-acc}
\end{figure}

\FloatBarrier
\section{PG19 Generational Examples}\label{sec:app:pg19-gen-sample}
In all generated samples, whitespaces are removed. Based on the prompt, presented in red, 256 bytes are generated and converted to a string. For conciseness, we show the start and end of the prompt and omit some content, which is denoted by an ellipsis. All prompts originate from books contained in the PG19 validation set \cite{pg19}.

\subsubsection*{2D-100K SSM-Transformer}
\begin{tcolorbox}[colback=lightgray!25, sharp corners]
\textcolor{red}{There is no sugar cane known anywhere to-day in th (...) number of otherwise remarkably distinct forms may be recognized some of which were illustrated in a previous publication, Bureau of Agriculture Bulletin No. 27, Citriculture in the Philippines, 1913, and referred to C. histrix with the statement that "some of these forms unquestionably will be recognized as subspecies on closer study, or possibly as separate species." Since then several plants of this type in the citrus collection assembl}ed and antimony in Aloeso are also combined. Those in the first fifteenth century likewise have been distinguished, but the species occurs in emergency and consists of two parts in the caterpillar which though often taken no leaves are next reduced. Specimens of other Granata have been found in a museum which was found generally at Manila . . and A. and occurred in Colorado and other small communities. Buddhisches, tompsing scientifically the time occupied in scientific research, have previously
\end{tcolorbox}

\begin{tcolorbox}[colback=lightgray!25, sharp corners]
\textcolor{red}{Dawson had often come in and out of the room durin (...) motion was required to alleviate the agony of fury that seized upon the Cagots at such times. In this desire for rapid movement, the attack resembled the Neapolitan tarantella; while in the mad deeds they performed during such attacks, they were not unlike the northern Berserker. In Béarn especially, those suffering from this madness were dreaded by the pure race; the Béarnais, going to cut their wooden clogs in the great forests that lay}around them, accumulated their old equipment, and spent their supplies under small vessels and towards the pillaginian regions which they would have drawn from it. But in so far as the Fairies were concerned on the matter, they were left to grant their reasons to the cautious Battery and his friends. The only alley of considerable importance, and whence too many of the adventurous scientific men, have appeared when they entered the school, or where the place has been called a masquerade and the choi
\end{tcolorbox}

\subsubsection*{2D-100K SSM-SSM}
\begin{tcolorbox}[colback=lightgray!25, sharp corners]
\textcolor{red}{He had an envelope in his starboard mitten, and, c (...) are forbidden crossing this property, under penalty of the law.' But land! I'd used that short-cut ever sence I'd been in Bayport--which was more'n a year--and old man Davidson and me was good friends, so I cal'lated the signs was intended for boys, and hove ahead without paying much attention to 'em. 'Course I knew that the old man--and, what was more important, the old lady--had gone abroad and that the son was expected down, but that d}idn't make it any good. The time was fast enough in the morning to launch up a fat pirate lord in a thresh-open carriage an' walk out to the dock, that can stand it on fifty yards with his head turned to look at the cruise. It's most lucky for a while. He's not at home this time. I guess he's gone to the pier pond. Said I was there and he says that that he can tell his 'and that's more the truth. To-morrow night I'll go down to see how Jim Buck works. I can't see how he's going. There's a chance for
\end{tcolorbox}

\begin{tcolorbox}[colback=lightgray!25, sharp corners]
\textcolor{red}{In a House of Commons that counted Pitt, Fox, Burk (...) usiness with his secretaries. Hundreds of times, probably, I have called him out of bed, and have, in short, seen him in every situation and in his most unreserved moments. As he knew I should not ask anything of him, and as he reposed so much confidence in me as to be persuaded that I should never use any information I might obtain from him for any unfair purpose, he talked freely before me of men and things, of actual, meditated, or ques}tionable, general matters, and of all matters that require the utmost collision. On one occasion the project proposed to place my position at the head of some five or six influential members of parliament on the line of steamships, or those who had a distinct presidential interest in the theatre, a misdemeanour and inducement making me the interest of the community towards the committee; the people of the country and statesmen of eminence at Lichfield could not be more maturely charged than I am than Mr. Gr
\end{tcolorbox}

% You can have as much text here as you want. The main body must be at most $8$ pages long.
% For the final version, one more page can be added.
% If you want, you can use an appendix like this one.  

% The $\mathtt{\backslash onecolumn}$ command above can be kept in place if you prefer a one-column appendix, or can be removed if you prefer a two-column appendix.  Apart from this possible change, the style (font size, spacing, margins, page numbering, etc.) should be kept the same as the main body.
%%%%%%%%%%%%%%%%%%%%%%%%%%%%%%%%%%%%%%%%%%%%%%%%%%%%%%%%%%%%%%%%%%%%%%%%%%%%%%%
%%%%%%%%%%%%%%%%%%%%%%%%%%%%%%%%%%%%%%%%%%%%%%%%%%%%%%%%%%%%%%%%%%%%%%%%%%%%%%%

\end{document}